\documentclass{article}

\usepackage[preprint]{corl_2026} % Uncomment for pre-prints (e.g., arxiv); This is like ``final'', but will remove the CORL footnote.

\title{CoReMatch: Recurrent Continuous Coordinate Refinement for Real-Time Robotic Feature Matching}

\author{
	Robert J. Wang$^{1}$, Charles Ling$^{2}$, Boyu Wang$^{2,3}$ \\
	$^{1}$Teledyne Technologies, 
	$^{2}$Western University, $^{3}$Vector Institute 
	\\
	\texttt{\{jwan563, charles.ling\}@uwo.ca, bwang@csd.uwo.ca} 
}

\usepackage{graphicx}
\usepackage{subcaption}
\usepackage{booktabs} % for professional tables
\usepackage{multirow}

\usepackage{tikz}
\usepackage{wrapfig}
\usepackage{amssymb}
\usepackage{makecell}
\usepackage{amsmath}
\usepackage{amssymb}

\begin{document}
\maketitle

%===============================================================================
\begin{abstract}
Feature matching underpins many robotic perception tasks, including visual localization, mapping, SLAM, and autonomous navigation. Existing semi-dense matchers predict confidence on fixed grids, limiting correspondences to predefined locations and requiring costly high-resolution grids for precise localization. We present CoReMatch (\textbf{Co}ordinate \textbf{Re}finement Match), a recurrent framework that refines correspondences directly in continuous coordinate space rather than predicting discrete confidence maps. CoReMatch progressively improves correspondence estimates by propagating information across iterations. At the coarse stage, a Deep Aggregated Transformer captures global context and iteratively updates correspondence coordinates, while a lightweight regression head further sharpens their localization. Experiments on indoor and outdoor benchmarks show that CoReMatch strikes a strong balance between accuracy and efficiency, outperforming Efficient LoFTR in accuracy while running more than $10\times$ faster on embedded edge hardware. These results demonstrate that recurrent coordinate refinement offers an efficient and scalable approach to real-time feature matching for robotic perception. Code is available at \href{https://github.com/edge-cv/CoReMatch}{edge-cv/CoReMatch}.

\end{abstract}

%===============================================================================

\section{Introduction}
\label{sec:intro}

Image feature matching establishes pixel-level correspondences between images and is a core component of robotic perception. It supports visual localization, mapping, pose estimation, and navigation in systems such as SLAM, VIO, and autonomous aerial robots. In real-time applications, matches must be computed under tight latency, memory, and power constraints, as unreliable correspondences can directly degrade trajectory estimation, map consistency, and control stability.
Classical matching pipelines rely on keypoint detection and descriptor matching, but often struggle with large viewpoint changes, motion blur, illumination variations, and low-texture regions. Learning-based methods improve robustness by combining dense visual features with global context. In particular, semi-dense matchers~\cite{sun2021loftr, yu2023adaptive, chen2022aspanformer, giang2023topicfm, ni2023pats, tang2022quadtree, wang2022matchformer, wang2024eloftr} replace explicit keypoint detection with coarse dense matching, followed by local refinement of high-confidence correspondences.

Despite their effectiveness, existing semi-dense matchers typically predict discrete confidence maps over spatial grids. Precise localization is therefore constrained by the resolution of the prediction grid, often requiring higher-resolution inputs to achieve accurate correspondences. However, the computational and memory costs of Transformer-based architectures increase substantially with input resolution. Although recent methods reduce theoretical FLOPs through linear or aggregated attention, practical runtime on embedded platforms can still be dominated by memory movement and feature aggregation. Consequently, modest increases in image resolution can lead to disproportionate latency growth. For example, as shown in Tab.~\ref{tab:efficiency_input}, Efficient LoFTR~\cite{wang2024eloftr} requires 311\,ms to process a $640\times640$ image pair, while increasing the resolution to $960\times960$ results in a $4.5\times$ runtime increase for only a $1.5\times$ increase in input resolution. At $1152\times1152$, it requires more than two seconds to process an image pair on an NVIDIA AGX Orin. This scaling behavior presents a significant challenge for real-time and on-device robotic perception, motivating more resolution-efficient approaches to correspondence estimation.

Recent dense matching methods~\cite{edstedt2023dkm, edstedt2024roma} obtain subpixel correspondences through continuous regression and iterative refinement. However, they commonly rely on correlation volumes and multi-scale warping, which introduce substantial computational and memory overhead. These limitations motivate a central question: \textbf{\textit{Can semi-dense matching produce accurate subpixel correspondences without high-resolution prediction grids or costly correlation-volume refinement?}}

\begin{figure}[t]
	\begin{center}
		\includegraphics[width=0.3\linewidth]{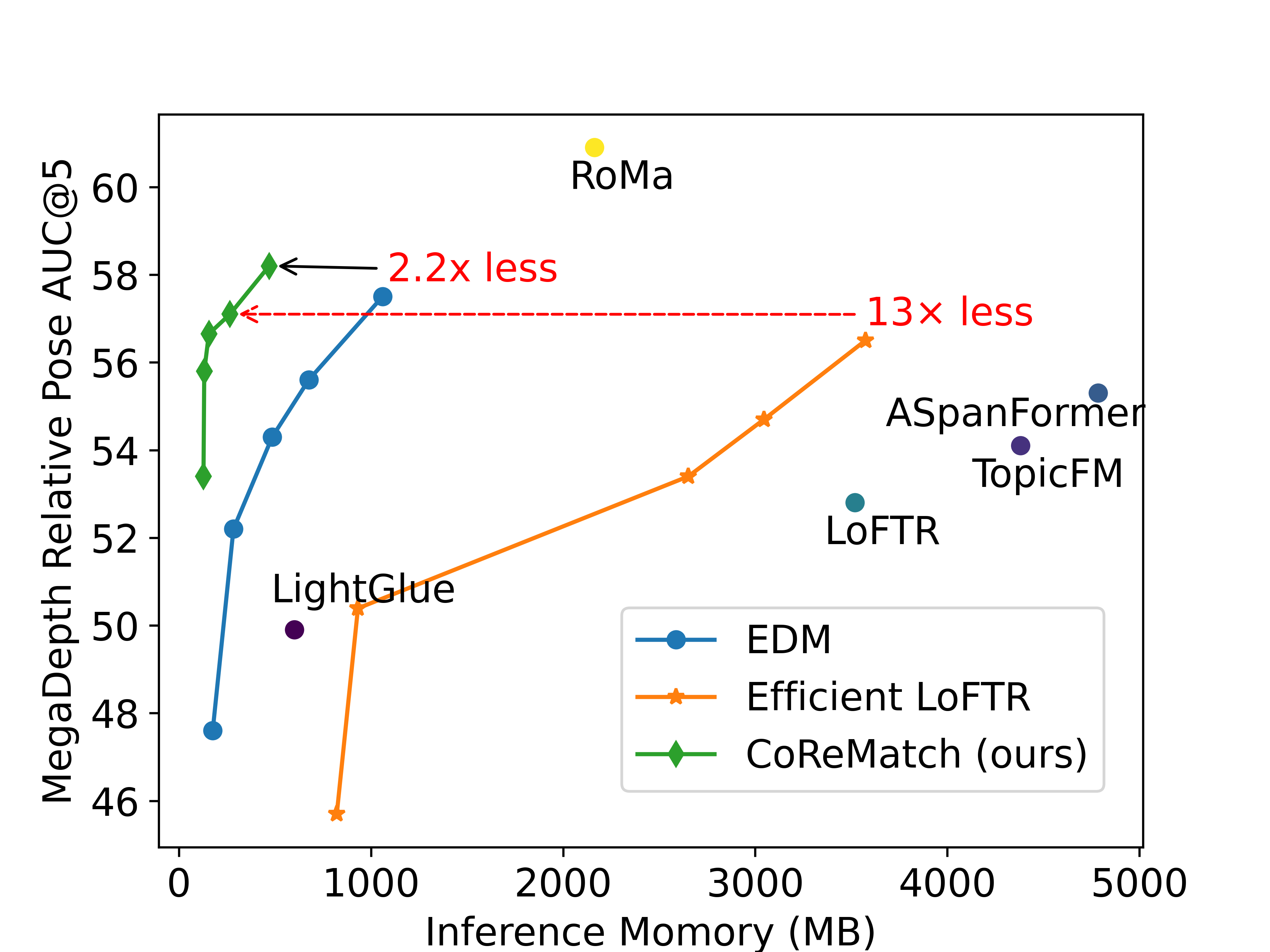}
		\includegraphics[width=0.3\linewidth]{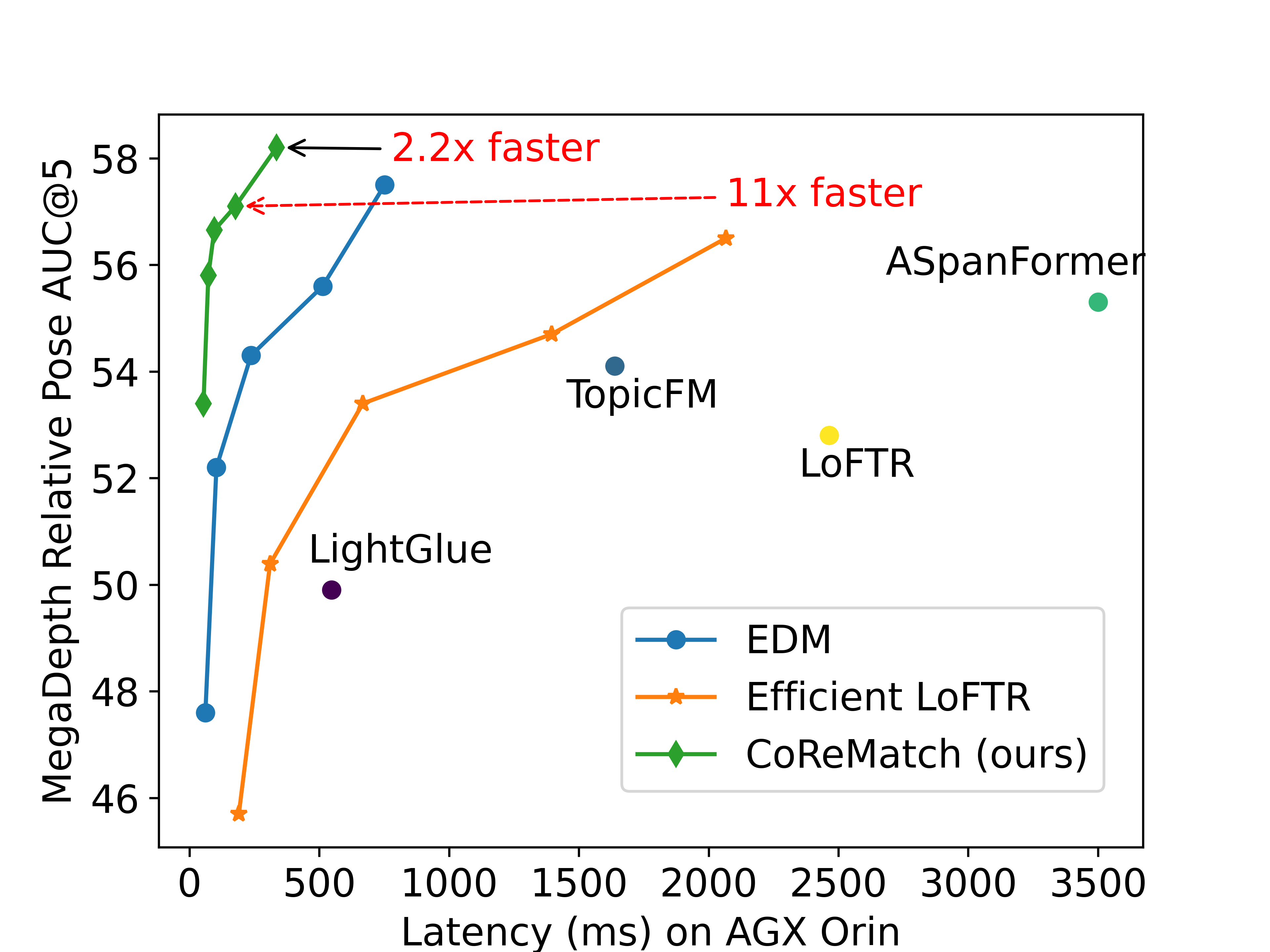}
		\includegraphics[width=0.3\linewidth]{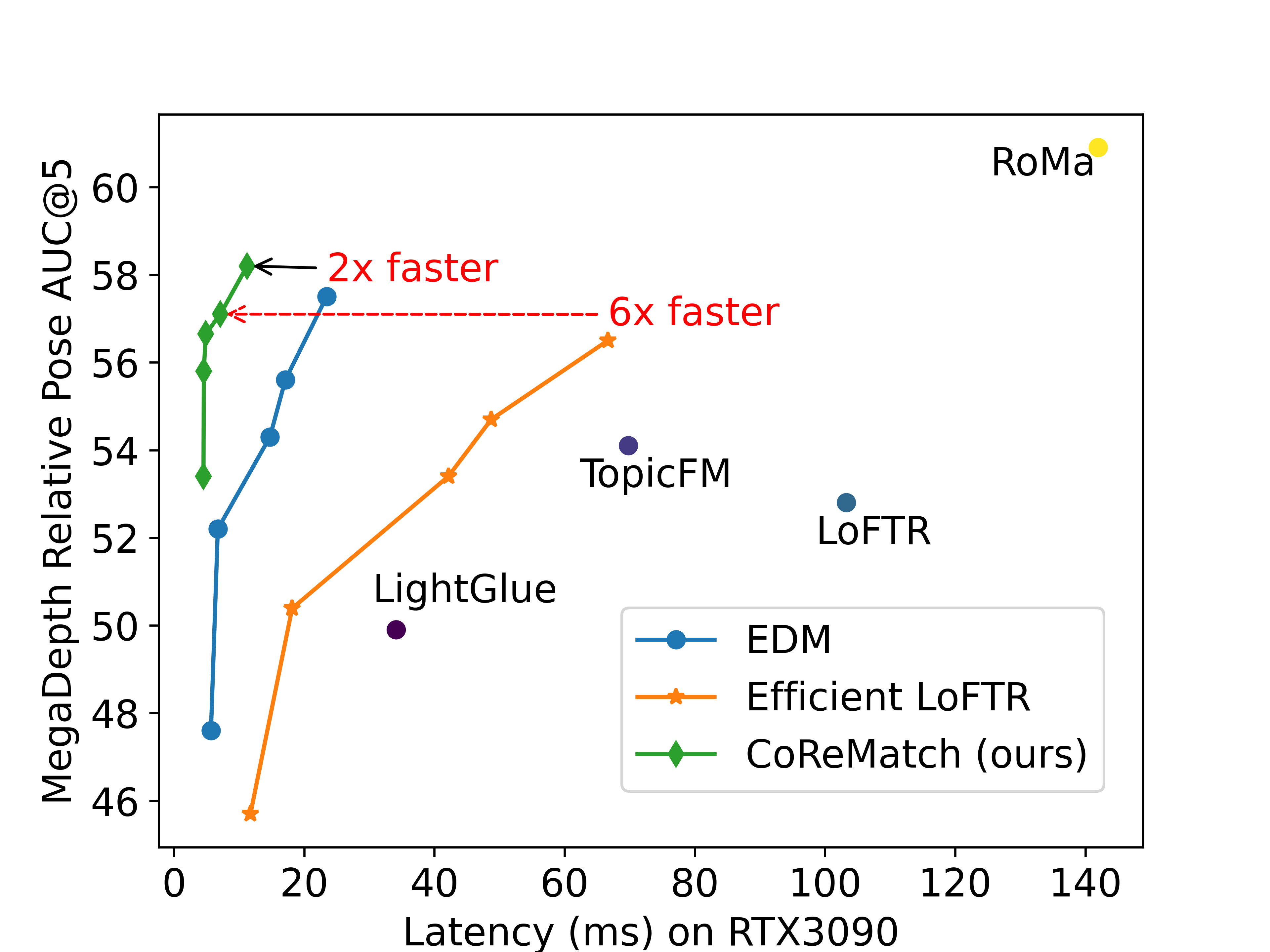}
	\end{center}
	\caption{
		\textbf{Memory--Speed--Accuracy Trade-off of State-of-the-Art Matchers on MegaDepth.}
		Memory usage and runtime are measured on NVIDIA AGX Orin and RTX 3090 GPUs using batch size 1 and FP16 precision.
		Higher values on the y-axis indicate better performance, while lower values on the x-axis indicate higher efficiency.
		Across both equal-resolution and equal-accuracy comparisons, CoReMatch consistently outperforms prior state-of-the-art methods.
		Additional quantitative results are provided in Table~\ref{tab:megadepth_and_scannet} and Table~\ref{tab:efficiency_input}.
	}
	
	\label{fig_megadepth}
\end{figure}

In this work, we answer this question with \emph{CoReMatch (Coordinate Refinement Match)}, an efficient semi-dense feature matching framework designed for real-time robotic perception. Unlike existing semi-dense methods that formulate matching as discrete confidence prediction over feature grids, CoReMatch reformulates correspondence estimation as a recurrent continuous coordinate refinement problem. CoReMatch directly refines correspondences in coordinate space through persistent hidden-state updates, significantly reducing computational and memory overhead while preserving accurate sub-pixel localization.
Specifically, CoReMatch adopts a two-stage recurrent refinement framework. At the coarse level, a Deep Aggregated Transformer (DAT) iteratively predicts and refines correspondences through direct coordinate updates while maintaining a persistent hidden state across iterations. This design enables efficient global geometric reasoning without explicit dense correlation-volume construction. The recurrent refinement process shares parameters across iterations, allowing the number of refinement steps to be dynamically adjusted at runtime based on hardware constraints or convergence behavior.
At the fine level, a lightweight convolutional network performs local dense regression within small full-resolution windows for efficient sub-pixel refinement. By combining recurrent coordinate-space refinement with sparse computation, CoReMatch bridges the gap between efficient semi-dense matching and accurate dense correspondence estimation.
Our contributions are summarized as follows:

\textbf{A recurrent continuous refinement framework for efficient semi-dense feature matching.} We formulate semi-dense matching as a recurrent continuous coordinate refinement problem, enabling accurate sub-pixel correspondence estimation without constructing a dense correlation volume. CoReMatch enables efficient geometric reasoning while reducing reliance on high-resolution feature maps. As a result, CoReMatch achieves higher accuracy at a lower input resolution (448$\times$448) than Efficient LoFTR at substantially higher resolutions.

\textbf{State-of-the-art accuracy--efficiency trade-offs with strong robotic generalization.}
We evaluate CoReMatch on multiple indoor and outdoor correspondence benchmarks, including MegaDepth \cite{li2018megadepth}, ScanNet \cite{dai2017scannet}, and HPatches \cite{balntas2017hpatches}. CoReMatch consistently outperforms state-of-the-art semi-dense methods while achieving over $10\times$ faster inference on embedded edge devices. Beyond benchmarks, CoReMatch improves robustness in downstream SLAM and VIO systems and demonstrates strong generalization to unseen environments and cross-modality RGB--infrared matching.

\section{Method}

\begin{figure}[t]
\begin{center}
\includegraphics[width=0.8\linewidth]{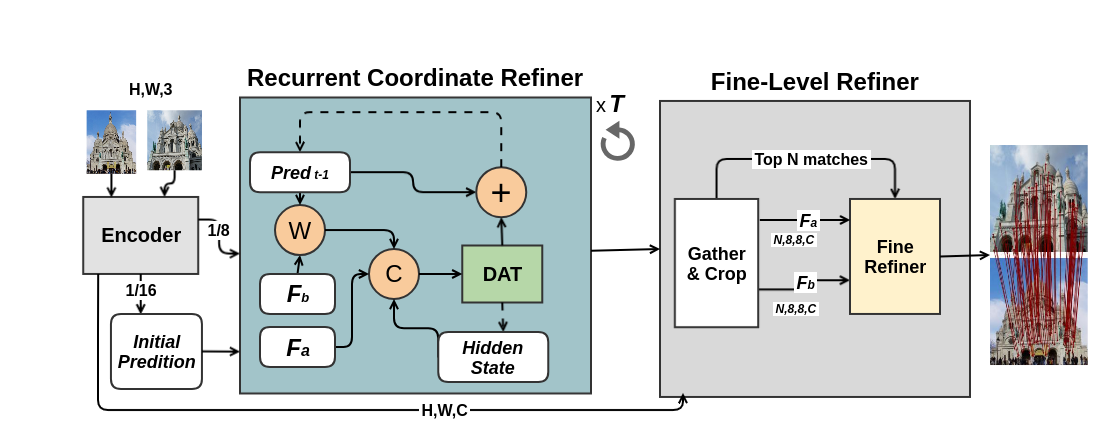}
\includegraphics[width=0.335\linewidth]{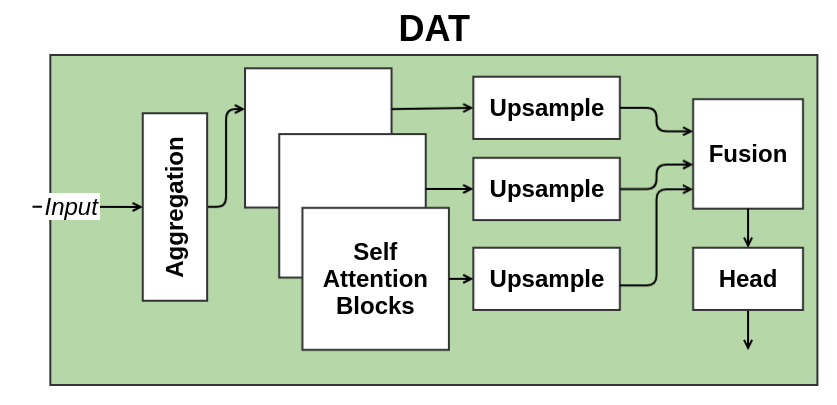}
\includegraphics[width=0.35\linewidth]{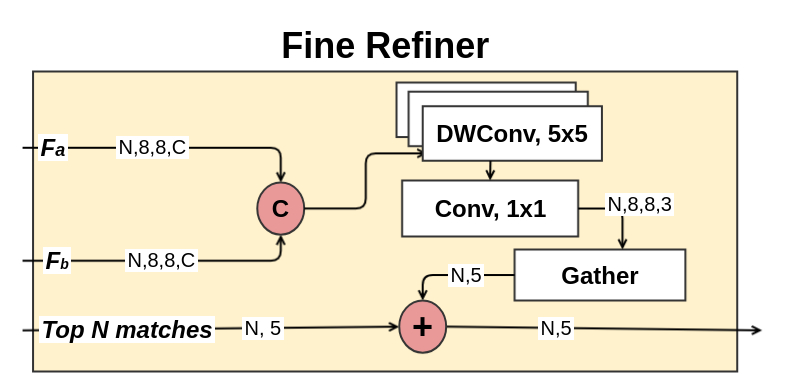}
\caption{\textbf{Overview of CoReMatch.}
	We propose a recurrent continuous coordinate refinement framework for semi-dense feature matching. Given an image pair, CoReMatch first predicts initial correspondences using low-resolution 1/16 feature maps. A recurrent coordinate refinement module then progressively updates the correspondences through iterative coordinate-space optimization using a Deep Aggregated Transformer (DAT) for global geometric reasoning. The resulting correspondences are further refined by a lightweight CNN that performs sub-pixel regression within compact local feature windows. This coarse-to-fine coordinate refinement design enables efficient and accurate matching for real-time robotic perception.}
	
\label{fig_corematch_arch}

\end{center}
\end{figure}

\subsection{Overview}

CoReMatch formulates feature matching as a recurrent continuous coordinate refinement problem for efficient semi-dense correspondence estimation. As shown in Fig.~\ref{fig_corematch_arch}, the framework follows a coarse-to-fine pipeline. DAT handles global semantic correspondence reasoning, while the fine-level module focuses solely on local geometric and confidence refinement. Overall, this recurrent coordinate refinement design combines the efficiency of semi-dense matching with accurate sub-pixel localization, making it suitable for real-time deployment on embedded robotic platforms.

\subsection{Coarse-Level Recurrent Coordinate Refinement}
\subsubsection{Correspondence Representation}

Given an image pair $(I_A, I_B)$, CoReMatch first extracts low-resolution feature maps 
$F_A \in \mathbb{R}^{H \times W \times C}$ and 
$F_B \in \mathbb{R}^{H \times W \times C}$ using a shared feature encoder. 
Unlike existing semi-dense methods that formulate matching as discrete confidence prediction over spatial grids, we represent correspondences directly in continuous coordinate space. Specifically, each query location is associated with a continuous correspondence coordinate
$\mathbf{x}_i = (u_i, v_i)$ in the target image together with a latent hidden state $\mathbf{h}_i$ that stores iterative geometric context during refinement.

We initialize a set of coarse query locations 
$\mathcal{Q} = \{\mathbf{q}_i\}_{i=1}^{N}$ 
on the source feature map using a regular low-resolution grid. Initial correspondences are obtained through global feature similarity matching on the low-resolution features, producing continuous target coordinates
$\mathcal{X}^{0} = \{\mathbf{x}_i^{0}\}_{i=1}^{N}$.
Rather than discretizing matches onto feature cells, CoReMatch maintains continuous coordinate estimates throughout the entire refinement process. This representation enables sub-pixel correspondence estimation even from low-resolution feature maps while avoiding the discretization artifacts commonly observed in grid-based semi-dense matching methods.

Each correspondence hypothesis is iteratively refined through recurrent coordinate updates:
\[
\mathbf{x}_i^{t+1} = \mathbf{x}_i^{t} + \Delta \mathbf{x}_i^{t},
\]
where $\Delta \mathbf{x}_i^{t}$ denotes the coordinate update predicted at iteration $t$.

\subsubsection{Recurrent Coordinate Update}

Starting from the initial dense correspondence field
$\mathcal{X}^{0} = {\mathbf{x}*i^{0}}*{i=1}^{N}$,
CoReMatch performs iterative refinement directly in continuous coordinate space. Unlike existing dense correspondence methods that rely on local correlation volumes and multi-scale warping, our refinement uses recurrent hidden-state propagation without explicit correlation-volume search, substantially reducing computational and memory overhead while retaining iterative geometric refinement.

At iteration $t$, each correspondence hypothesis consists of a source location $\mathbf{q}_i$, a target coordinate estimate $\mathbf{x}_i^{t}$, and a latent hidden state $\mathbf{h}_i^{t}$. Given the current coordinate estimate, target features are sampled from the feature map $F_B$ using differentiable bilinear interpolation:

$$
\mathbf{f}_{i}^{t} = \phi(F_B, \mathbf{x}_i^{t}),
$$

where $\phi(\cdot)$ denotes continuous feature sampling.

The sampled target feature, source feature, and previous hidden state are then processed by a recurrent Transformer update module:

$$
\mathbf{h}_i^{t+1} =
\mathcal{T}\left(
\mathbf{f}_{A,i},
\mathbf{f}_{i}^{t},
\mathbf{h}_i^{t}
\right),
$$

where $\mathcal{T}(\cdot)$ shares parameters across refinement iterations. The updated hidden representation is used to predict a continuous coordinate offset:

$$
\Delta \mathbf{x}_i^{t} = g(\mathbf{h}_i^{t+1}),
$$

followed by coordinate refinement:

$$
\mathbf{x}_i^{t+1} = \mathbf{x}_i^{t} + \Delta \mathbf{x}_i^{t}.
$$

\begin{figure}[ht]
	\centering
	\begin{minipage}[b]{0.48\textwidth}
		\centering
		\includegraphics[width=0.8\textwidth]{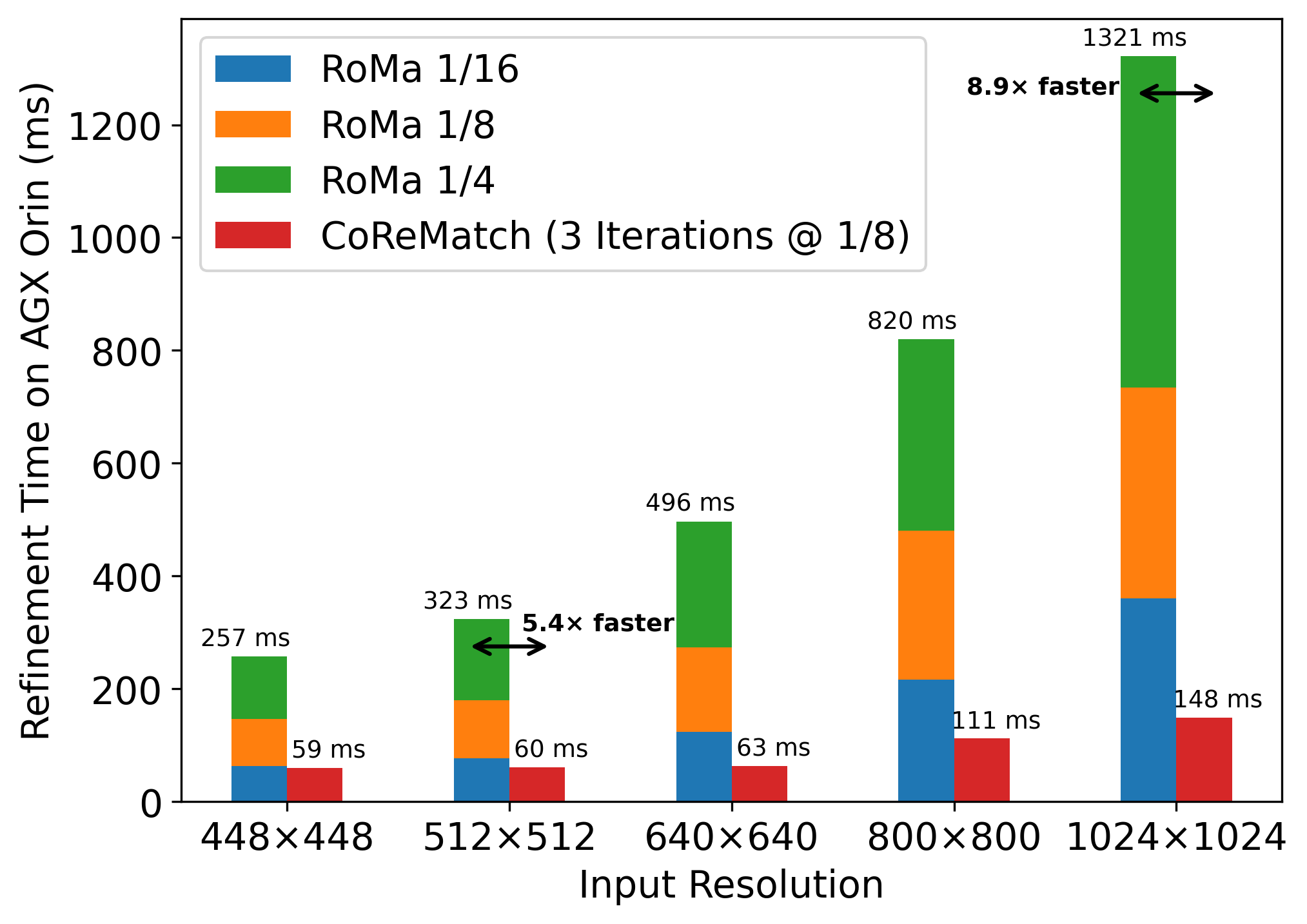}
		\caption{
			Runtime breakdown of refinement mechanisms.
			Compared with RoMa-style multi-scale correlation-volume refinement at 1/16, 1/8, and 1/4 resolutions, CoReMatch performs three recurrent refinement iterations entirely at 1/8 resolution, achieving over $5\times$ faster runtime at $512\times512$ and over $8\times$ faster runtime at $1024\times1024$.
		}
		\label{fig_speed_cfiner}
	\end{minipage}
	\hfill
	\begin{minipage}[b]{0.48\textwidth}
		\centering
		\includegraphics[width=0.8\textwidth]{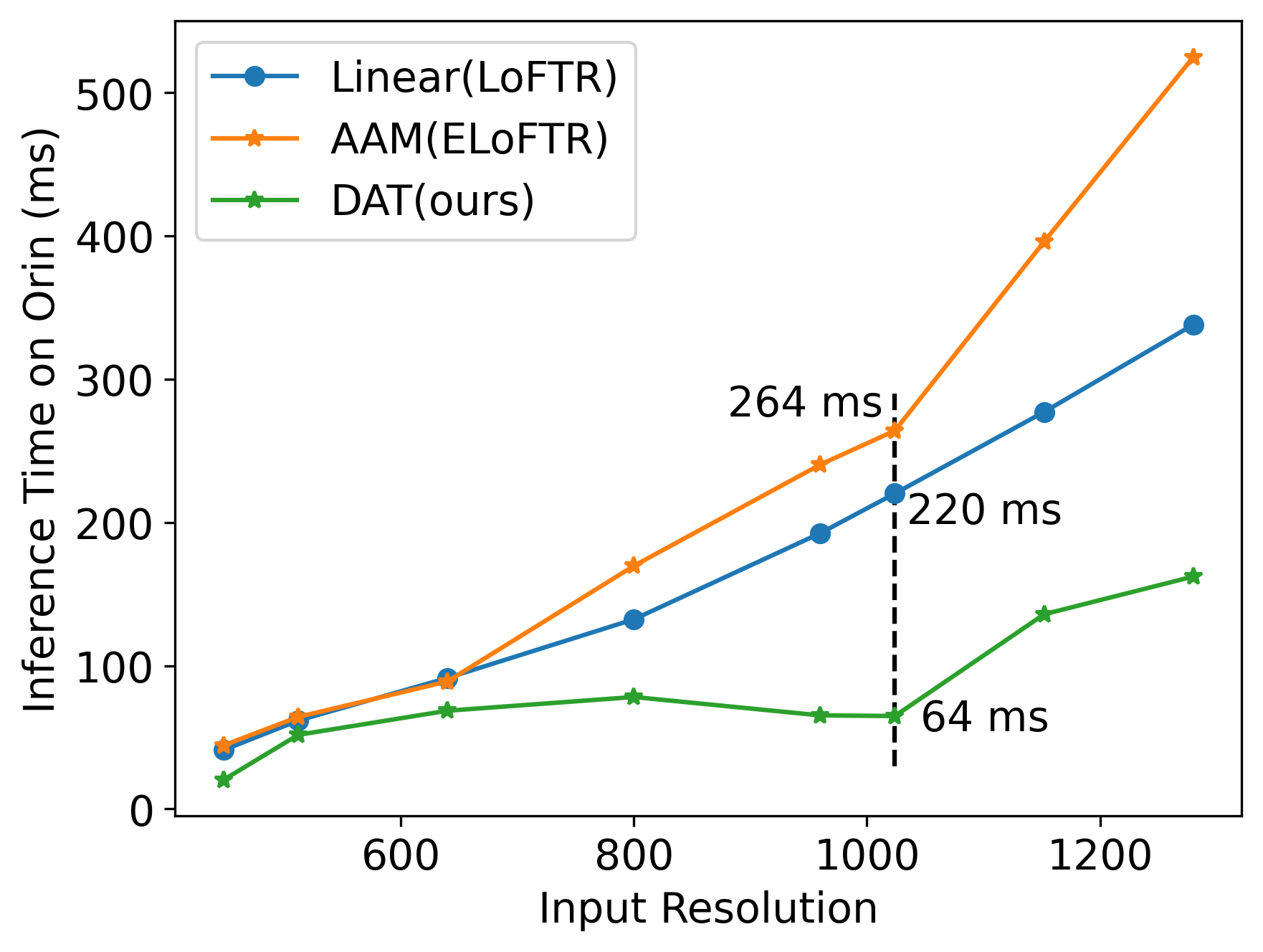}
		\caption{
			Inference time of different attention mechanisms on NVIDIA AGX Orin across varying input resolutions. The proposed Deep Aggregated Transformer (DAT) achieves over $4\times$ lower inference time than the AAM used in Efficient LoFTR at $1024\times1024$ resolution.
		}
		
		\label{fig_speed_attns}
	\end{minipage}
\end{figure}

The hidden state $\mathbf{h}_i^{t}$ is recurrently updated and propagated across iterations, enabling the model to accumulate geometric reasoning over time without constructing dense correlation volumes or performing multi-scale warping. As shown in Fig.~\ref{fig_speed_cfiner}, three recurrent iterations at the $1/8$ feature scale are over $5\times$ faster than RoMa-style multi-scale refinement at $1/16$, $1/8$, and $1/4$ scales for $512\times512$ inputs, and over $8\times$ faster for $1024\times1024$ inputs.

\subsubsection{Deep Aggregated Transformer}

Attention plays a central role in Transformer-based feature matching by enabling global geometric reasoning. However, standard attention has quadratic complexity with respect to the number of tokens, making it prohibitively expensive for dense correspondence estimation at high resolutions. To improve efficiency, LoFTR adopts linear attention, while Efficient LoFTR introduces an Aggregated Attention Module (AAM), which computes attention over a compact set of aggregated tokens.
Although AAM significantly reduces theoretical FLOPs, its practical efficiency on embedded platforms remains limited by repeated token aggregation and feature reconstruction operations across attention layers. These operations introduce substantial memory movement and kernel-launch overhead, which increasingly dominate runtime at high resolutions. As shown in Fig.~\ref{fig_speed_attns}, the inference time of AAM grows rapidly for resolutions larger than $640\times640$, limiting its scalability for real-time robotic perception.

To address this limitation, we propose a \emph{Deep Aggregated Transformer} (DAT), a memory-efficient architecture designed for recurrent coordinate refinement. Unlike Efficient LoFTR, DAT performs token aggregation only once at the Transformer input. The aggregated tokens are then processed by a deep Transformer operating entirely at the compact resolution, followed by a lightweight feature fusion block that combines intermediate and final Transformer features to recover spatial representations.
This deep-and-narrow design substantially reduces intermediate memory traffic while preserving strong global reasoning capability. By avoiding repeated aggregation and reconstruction operations across attention layers, DAT achieves significantly better scalability on embedded platforms. As shown in Fig.~\ref{fig_speed_attns}, DAT achieves over $4\times$ lower inference time than AAM at $1024\times1024$ resolution.

\subsection{Fine-Level Refinement}

Existing semi-dense matching methods such as \textit{LoFTR} and \textit{Efficient LoFTR} perform fine-level refinement by computing local correlations within cropped feature patches around coarse correspondences. However, refinement is typically formulated as discrete confidence prediction over local feature grids, which inherently limits sub-pixel localization accuracy.

To address this limitation, we reformulate fine-level refinement as a local continuous regression problem. Specifically, we select the top-$N$ coarse correspondences according to matching confidence for fine-level refinement. For each selected correspondence, an $8\times8$ local window is extracted from the fine-level feature maps, and the network directly predicts continuous coordinate offsets within the window. This formulation enables sub-pixel refinement directly in coordinate space without relying on discretized local matching grids.
Since semantic correspondence estimation is largely resolved at the coarse stage, the fine-level module focuses mainly on local geometric refinement. As illustrated in Fig.~\ref{fig_corematch_arch}, the refinement network consists of three depthwise separable convolutional layers followed by a lightweight $1\times1$ prediction head. The lightweight design introduces minimal computational overhead while preserving accurate sub-pixel localization, making it well suited for real-time robotic perception on embedded platforms.

\subsection{Feature Representation}

CoReMatch is backbone-agnostic and does not depend on a specific feature encoder. For real-time deployment, we use a lightweight PeleeNet-based backbone~\cite{huang2017densely,wang2018pelee}, providing a strong speed--accuracy trade-off on embedded platforms. To evaluate generalization, we also integrate frozen DINOv2~\cite{dosovitskiy2020image,oquab2023dinov2} features into the same refinement framework. The DINOv2-based variants achieve stronger performance than the lightweight CNN backbone on MegaDepth while substantially improving robustness across unseen domains and sensing modalities without fine-tuning, including indoor scenes, autonomous driving, UAV imagery, and RGB--infrared matching.

We further provide scalable variants under the same recurrent refinement framework. CoReMatch-S employs a lightweight CNN backbone with a single refinement iteration for maximum efficiency on embedded platforms. CoReMatch-M adopts a frozen DINOv2 ViT-S backbone with two refinement iterations for improved robustness and cross-domain generalization. CoReMatch-L further scales the framework using a DINOv2 ViT-L backbone with three refinement iterations for higher matching accuracy in challenging large-scale environments. All variants share the same recurrent coordinate refinement formulation, demonstrating that CoReMatch scales effectively from lightweight embedded deployment to high-capacity foundation-model-based robotic perception systems.

\subsection{Training Objectives}

CoReMatch is trained in a two-stage strategy. We first train the coarse-level matcher, after which its weights are frozen and the fine-level refiner is trained. Since both the coarse matcher and the fine-level refiner predict dense correspondence fields along with confidence scores, they share the same loss formulation.

The overall loss consists of a regression term and a confidence term:
\begin{equation}
	\mathcal{L}
	=
	\mathcal{L}_{\mathrm{warp}}
	\left(
	\hat{W}_l^{\mathcal{A} \rightarrow \mathcal{B}}
	\right)
	+
	\lambda
	\mathcal{L}_{\mathrm{conf}}
	\left(
	\hat{p}_l^{\mathcal{A} \rightarrow \mathcal{B}}
	\right),
\end{equation}
where $\lambda = 0.01$ balances the two terms, following DKM and RoMa.

The confidence loss $\mathcal{L}_{\mathrm{conf}}$ is implemented using binary cross-entropy.
For the regression loss $\mathcal{L}_{\mathrm{warp}}$, we adopt the generalized Charbonnier loss~\cite{barron2019general}:
\begin{equation}
	\mathcal{L}_{\mathrm{warp}}(y,p)
	=
	\sum_{i=1}^{N}
	\left[
	\left(
	\frac{y_i - p_i}{c}
	\right)^2
	+ 1
	\right]^{\frac{1}{2}},
\end{equation}
where $c = 0.001$ controls the width of the quadratic region near zero.

\section{Experiments}
\label{sec:experiments}

We evaluate CoReMatch on homography and relative pose estimation across diverse indoor and outdoor benchmarks. To assess downstream performance, we further evaluate visual odometry on four datasets. All evaluations use a single model trained exclusively on MegaDepth, enabling a direct assessment of zero-shot generalization. Finally, we conduct extensive ablation studies to analyze key architectural and training configurations.

\subsection{Training Details}

All models are trained on MegaDepth using the standard split \cite{sun2021loftr, edstedt2024roma}. Training is performed on three NVIDIA A6000 GPUs with a batch size of 24 using AdamW \cite{loshchilov2017decoupled}. The learning rate is set to $1\times10^{-4}$ for the backbone and $3\times10^{-4}$ for the matcher and refiner. Training uses a fixed resolution of $512\times512$. Ablation models are trained for 30 epochs, while final models are trained for 100 epochs.

\subsection{Relative Pose Estimation}

\noindent\textbf{Dataset and Evaluation Protocol.}
We evaluate relative pose estimation on MegaDepth \cite{li2018megadepth} and ScanNet \cite{dai2017scannet}. Predicted correspondences are rescaled to the original image resolution following DKM and RoMa. Relative poses are estimated from matches and evaluated using the maximum angular error between rotation and translation. We report AUC at 5°, 10°, and 20°. For LightGlue, runtime includes SuperPoint extraction (2048 keypoints).

\begin{table*}[htbp]
		\footnotesize
	\centering
	
\caption{\textbf{Relative Pose Estimation Results.}
All methods are trained on MegaDepth and evaluated on both the MegaDepth and ScanNet datasets.
	We report the AUC of relative pose error at multiple thresholds and average inference time (ms).
	Inference time is measured on an AGX Orin with batch size 1 and FP16 precision.
	CoReMatch-S outperforms ELoFTR by a large margin while being more than $11\times$ faster on MegaDepth.}

	\begin{tabular}{llcccc}
		\toprule
		\small
		\multirow{2}*{Category} & \multirow{2}*{Method}                                                & \multicolumn{2}{c}{MegaDepth} &  \multicolumn{2}{c}{ScanNet}                                                                          \\
		\cmidrule(lr){3-4}
		\cmidrule(lr){5-6}
		& & AUC@$5^{\circ}$/$10^{\circ}$ /$20^{\circ}$&Time& AUC@$5^{\circ}$/$10^{\circ}$ /$20^{\circ}$  &   Time
		
		\\\midrule
		\multirow{2}*{Sparse}
		& SP \cite{detone2018superpoint} + SG \cite{sarlin2020superglue}       & 49.7 / 67.1 / 80.6 &  -  & 16.2 / 32.8 / 49.7      &     -                     \\
		& SP \cite{detone2018superpoint} + LG \cite{lindenberger2023lightglue} & 49.9 / 67.0 / 80.1 &  548 & 14.8 / 30.8 / 47.5            &461            
		\\\midrule
		\multirow{3}*{Dense}
		& DKM \cite{edstedt2023dkm} & 60.4 / 74.9 / 85.1  & 7110   & 26.6 / 47.1 / 64.2  &  1418  \\
		& ROMA \cite{edstedt2024roma} & 60.9 / 75.6 / 85.6 & 4964 & 28.9 / 50.4 / 68.3&4964 \\
		\midrule
		\multirow{7}*{Semi-Dense}
		& LoFTR \cite{sun2021loftr} & 52.8 / 69.2 / 81.2 & 2465 & 16.9 / 33.6 / 50.6&  437 \\

		& QuadTree \cite{tang2022quadtree}& 54.6 / 70.5 / 82.2&-& 19.0 / 37.3 / 53.5 &     -     \\
		& ASpanFormer \cite{chen2022aspanformer}& 55.3 / 71.5 / 83.1 &3501& 19.6 / 37.7 / 54.4 &    1078      \\
		& TopicFM \cite{giang2023topicfm} & 54.1 / 70.1 / 81.6 & 1639& 17.3 / 35.5 / 50.9 &      260        \\
		& EfficientLoFTR \cite{wang2024eloftr} & 56.4 / 72.2 / 83.5   & 2067  & 19.2 / 37.0 / 53.6       &       311          \\
		& EDM~\cite{Li_2025_ICCV} & 57.5 / 73.2 / 84.2   &  752  & 19.8 / 37.5 / 54.4    &     104 \\
		& \textbf{CoReMatch-S}(Ours) & 57.4 / 72.8 / 83.7   &  \textbf{177}  & 22.8 / 42.7 / 60.6    &   \textbf{95} \\
		& \textbf{CoReMatch-M}(Ours) & 58.2 / 73.3 / 84.0 &  335 & 24.4 / 44.2 / 61.7&  161 \\

		& \textbf{CoReMatch-L}(Ours) & \textbf{59.5} / \textbf{74.4} / \textbf{84.8}   & 639  & \textbf{27.9} / \textbf{49.1} / \textbf{66.6}      & 366 \\
		\bottomrule
	\end{tabular}
	\label{tab:megadepth_and_scannet}
\end{table*}

\subsection{Homography Estimation}

 We evaluate homography estimation on the HPatches dataset \cite{balntas2017hpatches}, which consists of planar image sequences captured under varying viewpoints and illumination conditions. Predicted correspondences are used to estimate homographies via RANSAC, and accuracy is measured by the mean reprojection error of corner points. We report the Area Under the Curve (AUC) at 3, 5, and 10 pixel thresholds. For fair comparison, all methods share the same RANSAC settings, and for semi-dense approaches only the top 1000 predicted matches are retained, following LoFTR.
 As shown in Tab.~\ref{tab_hpatches}, our method demonstrates strong performance, significantly outperforming existing semi-dense approaches.

\begin{table}[ht]
	\centering
	\small
	\begin{minipage}{0.45\textwidth}
		\centering
\caption{\textbf{Homography Estimation on HPatches.}
	All methods are evaluated with models trained on MegaDepth to assess cross-dataset generalization.
	CoReMatch outperforms both sparse and semi-dense methods, and surpasses the state-of-the-art dense method RoMa.}

		\begin{tabular}{lccc}
	\toprule
	
	{\multirow{2}{*}{Method}}
	& \multicolumn{3}{c}{AUC}\\
	\cmidrule(r){2-4} 
	
	& @3px & @5px & @10px \\
	\midrule
	SuperGlue  &53.9 & 68.3 & 81.7 \\
	
	\midrule
	
	LoFTR&  65.9 & 75.6 & 84.6\\
	\small{DRC-Net} 
	& 50.6 & 56.2 & 68.3  \\
	ELoFTR  &   66.5 & 76.4 & 85.5   \\
		EDM  &   68.5 & 78.1 & 86.6   \\
	
	\textbf{CoReMatch-S }&69.5 & 79.1 & 87.7\\
	\textbf{CoReMatch-M }&70.3 & 79.7 & 88.0\\
	\textbf{CoReMatch-L }&\textbf{71.2} & \textbf{80.6} & \textbf{88.8}\\
	\midrule
	DKM & \textbf{71.2}&\textbf{80.4}&\textbf{88.4} \\
	RoMa  & 70.6 & 79.9 & 88.2 \\
	
	\bottomrule
\end{tabular}
		\label{tab_hpatches}
	\end{minipage}
	\hfill 
	\begin{minipage}{0.48\textwidth}
		\centering
		\small
		
		\caption{\textbf{Ablation Studies on MegaDepth.} Each component in CoReMatch improves AUC@10 while maintaining efficient inference.}
		
		\begin{tabular}{lcc}
			\toprule
			Setup & Time & AUC \\
			\midrule
			
			Baseline: ELoFTR-512 & 191.4 & 62.8 \\
			\midrule
			\textbf{Coarse Matching Module}&&\\
			RoMa-style & 408.3 & 68.5 \\
			
			RoMa-style (w/o correlation volume) & 295.9 & 67.3 \\
			Ours (1 iteration) & 158.6 & 68.9 \\
			Ours (2 iterations) & 178.7 & 69.7 \\
			\midrule
			
			\textbf{Fine-Level Refinement}&& \\
			Top-1 confidence (max)& 98.9 & 69.1 \\
			Mutual nearest neighbor (MNN) & 104.4 & 69.5 \\
			\midrule
			
			\textbf{Feature Encoder} &&\\
			Our CNN encoder & 72.9 & 70.0 \\
			DINOv2 ViT-S encoder & 144.2 & 71.1 \\
			\bottomrule
		\end{tabular}
		
		\label{tab_ablation}
	\end{minipage}
\end{table}

\subsection{Visual Odometry Evaluation}
We integrate CoReMatch as the feature-matching frontend within PySLAM and evaluate its performance against LoFTR, EfficientLoFTR (ELoFTR), and CoReMatch-S using a unified backend pipeline. With identical system configurations across all methods, we report unaligned Absolute Trajectory Error (ATE RMSE, in meters) on challenging sequences from four diverse datasets:TUM~\cite{sturm12iros} (\textit{fr1/desk}), EuRoC~\cite{Burri25012016} (\textit{V1\_03\_difficult}), KITTI~\cite{geiger2013vision} (\textit{06}), and TartanAir~\cite{wang2020tartanair} (\textit{abandonedfactory/P001}). As presented in Tab.~\ref{tab:vio}. CoReMatch achieves the lowest ATE on three of the four sequences while remaining highly competitive on TUM. These results demonstrate that CoReMatch translates effectively to downstream visual odometry while maintaining strong trajectory accuracy.

\begin{table}[t]
	\centering
	\footnotesize
	\setlength{\tabcolsep}{5pt}
	\caption{\textbf{Visual Odometry Evaluation.} We report unaligned ATE RMSE ($\downarrow$) on selected sequences. Best results are in bold. 		CoReMatch achieves the lowest ATE on 3/4 evaluated sequences and remains
			comparable to both baselines on TUM.}
\label{tab:vio}
\begin{tabular}{lcccc}
	\toprule
	Dataset & LoFTR & ELoFTR & CoReMatch & $\Delta$ vs. ELoFTR\\
	\midrule
	TUM~\cite{sturm12iros}  & \textbf{0.5931} & 0.5932 & 0.5934 & -0.03\% \\
	EuRoC~\cite{Burri25012016}  & 2.3859 & 2.2769 & \textbf{2.0673}& \textbf{+9.2\%}  \\
	KITTI~\cite{geiger2013vision}  & 10.5482 & 10.1711 & \textbf{6.9788}& \textbf{+31.4\%}  \\
	TartanAir~\cite{wang2020tartanair}  & 35.0914 & 35.4660 & \textbf{33.7973}& \textbf{+4.7\%} \\
	\bottomrule
\end{tabular}
\end{table}

\begin{table}[ht]
	\centering
\caption{\textbf{Comparison with Alternative Attention Mechanisms.} SCT denotes Simple Compact Transformer, while PTA denotes Proxy-Token Attention. Inference time is measured on an RTX 3090 with batch size 1 and FP16 precision.}

	\label{tab:sota2}
	\begin{tabular}{lcccc}
		\toprule
			&  DAT & WAFT~\cite{wang2026waft} & SCT   & PTA~\cite{wang2025ptdit}\\
		
		\midrule
		
Reprojection AUC@3px (\%) $\uparrow$ & 95.5 & 67.7 & 89.1 & 92.2 \\
Inference Time (ms) $\downarrow$     & 4.6  & 19.3 & 4.1  & 4.7  \\
		\bottomrule
	\end{tabular}
\end{table}

\subsection{Ablation Studies}

\subsubsection{Contribution of Each Component}

We conduct ablation studies to evaluate the contribution of each component in CoReMatch. The results are summarized in Tab.~\ref{tab_ablation}. Replacing ELoFTR's discrete coarse matcher with a RoMa-style regression-based matcher substantially improves matching accuracy, but at a significant cost in inference efficiency. In contrast, our proposed coarse matcher further improves accuracy while maintaining efficient inference, and the fine-level refiner provides an additional improvement in accuracy with only a modest computational overhead.

\subsubsection{Comparison with Alternative Attention Mechanisms}
In Tab.~\ref{tab:sota2}, we evaluate DAT against a WAFT-style~\cite{wang2026waft} structure and two token-reduction techniques: Simple Compact Transformer (SCT) and Proxy-Token Attention~\cite{wang2025ptdit} (PTA). DAT achieves the highest reprojection AUC@3px of 95.5\%, substantially outperforming SCT (89.1\%) and PTA (92.2\%). Meanwhile, DAT maintains competitive efficiency, with an inference time of 4.6 ms, compared with 4.1 ms for SCT and 4.7 ms for PTA.

\subsubsection{Ablation of Refinement Iterations and Top-Confidence Ratio}

As shown in Tab.~\ref{tab:iter}, the first two refinement iterations yield substantial performance gains, while the third iteration provides only a marginal improvement. Regarding the top-$K$ ratio, performance remains relatively stable once the ratio exceeds 20\%.

\begin{table}[t]
	\centering
	\setlength{\tabcolsep}{5pt}

\caption{\textbf{Refinement Ablation.} Results on the MegaDepth dataset with different refinement iterations and top-confidence ratios. All results use the same model without retraining. Inference time is measured on an AGX Orin with batch size 1 and FP16 precision.}

	\label{tab:iter}
	\begin{tabular}{lcccccccc}
		\toprule
		&\multicolumn{4}{c}{Iteration} &  \multicolumn{4}{c}{Top-$K$ Ratio (\%)}                                                                    \\
		\cmidrule(lr){2-5}\cmidrule(lr){6-9}
		& 0 & 1 & 2 & 3  & 10& 20 & 35 & 50\\
		\midrule
Reprojection AUC@3px (\%) $\uparrow$ & 29.2 & 92.8 & 95.5 & 95.9 & -- & -- & -- & -- \\
Relative Pose AUC@10 (\%)  $\uparrow$     & --   & 71.0 & 71.6 & 71.7 & 67.4 & 70.5 & 71.6 & 71.6 \\
Inference Time (ms) $\downarrow$& --   & 72.9 & 86.5 & 98.1 & 83.4 & 84.2 & 86.5 & 87.5 \\
		\bottomrule
	\end{tabular}
\end{table}

\section{Limitations and Future Work}

While CoReMatch achieves a strong accuracy--efficiency trade-off for real-time feature matching, several limitations remain and suggest promising directions for future work.

\textbf{Hardware optimization.}
CoReMatch is primarily optimized for GPU-based platforms. Although its design reduces memory movement and refinement overhead, further optimization is needed for deployment on mobile and ultra-low-power platforms. 

\textbf{Training data and generalization.}
Although CoReMatch generalizes well across indoor, driving, UAV, and RGB--IR scenarios, its training still relies primarily on MegaDepth supervision. Larger-scale and more diverse training data could further improve robustness in challenging real-world conditions. Exploring more diverse correspondence supervision and self-supervised training could further reduce the dependence on curated geometric datasets.

\textbf{Geometry-aware refinement.}
CoReMatch performs correspondence refinement through learned coordinate updates without explicitly incorporating geometric constraints. Integrating geometric priors, pose constraints, or temporal consistency could further improve performance in SLAM and visual odometry, particularly in challenging sequences with ambiguous or repetitive visual structures, where geometric consistency across frames becomes especially important.

\section{Conclusion}

We introduced CoReMatch, a semi-dense feature matching framework that replaces discrete grid-based confidence prediction with continuous coordinate-space refinement. CoReMatch combines a Deep Aggregated Transformer for global correspondence reasoning with a lightweight local regression module for precise refinement, achieving sub-pixel accuracy with substantially lower computational and memory costs than prior semi-dense methods.

Across standard benchmarks, CoReMatch achieves a strong accuracy--efficiency trade-off, consistently outperforming prior semi-dense methods while running over $10\times$ faster on embedded hardware. Its integration with frozen foundation-model features further improves robustness across unseen indoor, driving, UAV, and visible--thermal scenarios without fine-tuning. These results demonstrate that accurate geometric correspondence can be achieved without high-resolution prediction grids or explicit correlation-volume construction, providing an efficient alternative to prevailing designs for semi-dense matching.

\section*{Acknowledgments}
This work was supported by the Natural Sciences and Engineering Research Council of Canada (NSERC), Discovery Grants program.

%===============================================================================

%===============================================================================

\bibliography{main}  % .bib
\newpage
\appendix
% \onecolumn

\section{Appendix: Additional Quantitative Results}

Tab.~\ref{tab:efficiency_input} compares CoReMatch with ELoFTR and EDM across input resolutions in terms of relative pose estimation accuracy (AUC), inference time, and GPU memory consumption. At comparable accuracy levels, CoReMatch achieves substantially higher efficiency, running $2.2\times$ faster than EDM and over $11\times$ faster than ELoFTR, while reducing GPU memory consumption by more than $13\times$ compared with ELoFTR.

CoReMatch is particularly effective at low-resolution inference, achieving comparable AUC at $448\times448$ to ELoFTR at $800\times800$. At the same $640\times640$ input resolution, CoReMatch substantially outperforms both ELoFTR and EDM in accuracy (56.7 vs.\ 51.0 AUC) while requiring less inference time and GPU memory. These results demonstrate a substantially improved accuracy--efficiency trade-off, making CoReMatch well suited for real-time robotic perception on resource-constrained platforms.

\begin{table}
	 \small
	\centering
\caption{\textbf{Efficiency Comparison Across Input Resolutions.}
	We compare ELoFTR, EDM, and CoReMatch in relative pose estimation accuracy (AUC), GPU memory usage, and inference time on the NVIDIA AGX Orin and RTX 3090.
	CoReMatch achieves a favorable trade-off between accuracy and efficiency across input resolutions.}
			
	\begin{tabular}{lccclc}
		\toprule
		\multirow{2}*{Method}       & \multirow{2}*{Resolution}        & \multicolumn{1}{c}{Pose Est. AUC}        & \multicolumn{2}{c}{Runtime (ms)} & \multirow{2}*{Memory (MB)}  \\
		\cmidrule(lr){3-3}
		\cmidrule(lr){4-5}
		&                              & @$5^{\circ}$/@$10^{\circ}$/@$20^{\circ}$ & Orin & 3090     &      \\
		\midrule
		\multirow{5}*{ELoFTR}   
& 512$\times$512 & 45.7/62.9/76.1  & 191.4 & 11.7 & 822    \\
& 640$\times$640 & 51.0/67.4/79.8  & 310.6 & 18.2 & 932    \\
& 800$\times$800 & 53.4/70.0/81.9  & 668.1 & 42.2 & 2652    \\
& 960$\times$960 & 54.7/70.7/82.4  & 1394.9 & 48.7 & 3046    \\
& 1152$\times$1152 & 56.4/72.2/83.5  & 2066.7 & 66.7 & 3574    \\
		\midrule
\multirow{5}*{EDM}   
& 512$\times$512 & 47.6/64.7/77.8  & 61.5 & 5.7 & 176    \\
& 640$\times$640 & 52.2/68.9/80.9  & 104.1 & 6.7 & 284    \\
& 800$\times$800 & 54.3/70.8/82.4  & 237.3 & 14.7 & 486    \\
& 960$\times$960 & 55.6/71.4/82.8  & 513.2 & 17.1 & 676    \\
& 1152$\times$1152 & 57.5/73.2/84.2  & 752.3 & 23.5 & 1060    \\
		\midrule
\multirow{5}*{CoReMatch-S}   
& 448$\times$448 & 53.4/69.8/81.6  & 54.2 & 4.5 & 128    \\
& 512$\times$512 & 55.8/71.6/83.0  & 72.9 & 4.6 & 132    \\
& 640$\times$640 & 56.7/72.3/83.4  & 95.8 & 6.4 & 156    \\
& 1024$\times$1024 & 57.4 / 72.8 / 83.7  & 177.1 & 7.1 & 266    \\
& 1152$\times$1152 & 57.3 / 72.6 / 83.6  & 243.4 & 12.2 & 322    \\
		\bottomrule
	\end{tabular}

	\label{tab:efficiency_input}
\end{table}

\section{Appendix: Qualitative Results}

\begin{figure}[h]
	\begin{center}
		\begin{small}
			\begin{tabular}{lccc}
				\toprule
			&	ELoFTR & EDM & Ours\\
				\midrule
				{\multirow{2}{*}{\rotatebox[origin=c]{90}{MegaDepth}}} &
				\includegraphics[width=0.3\linewidth]{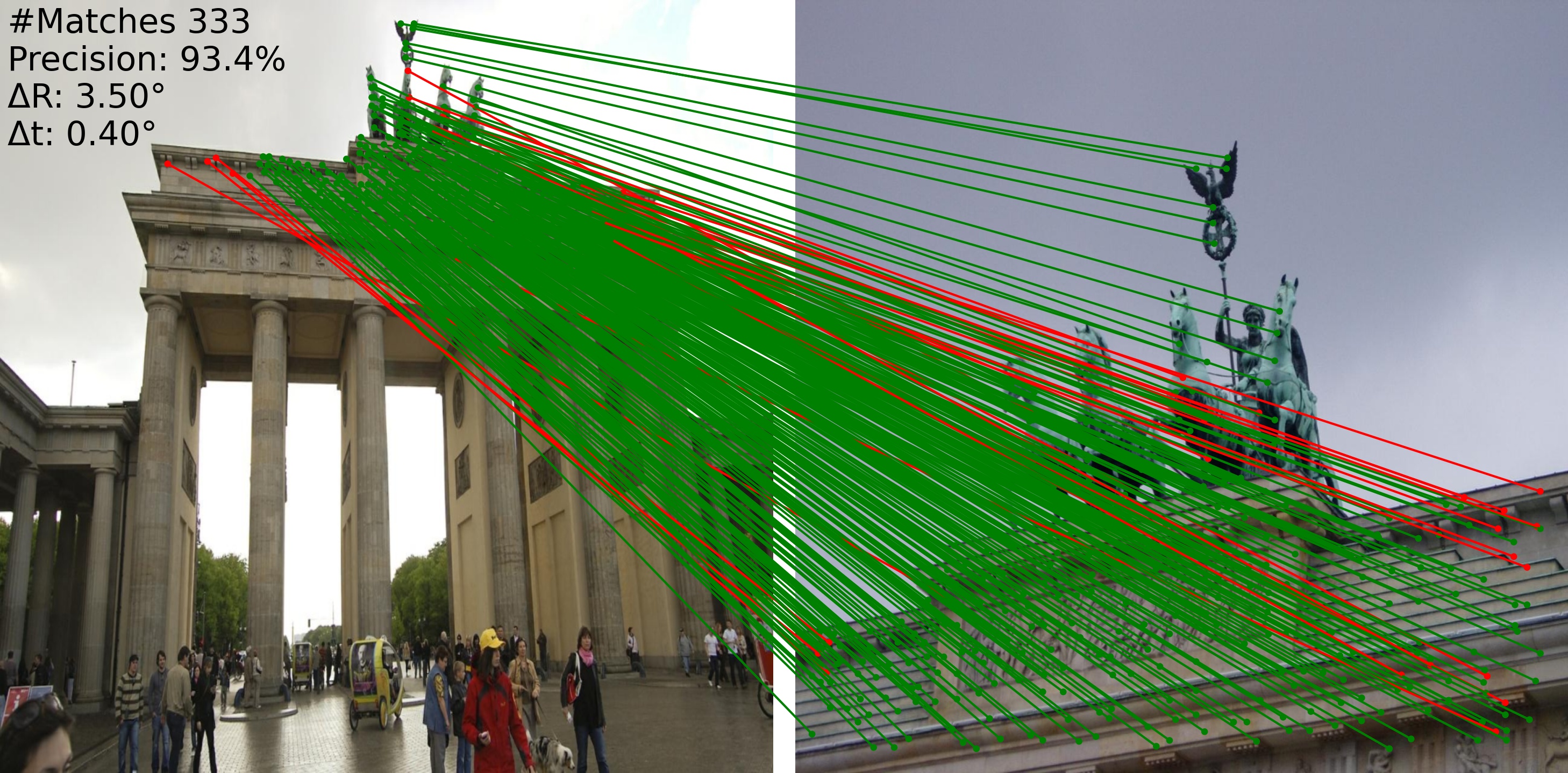}
				& \includegraphics[width=0.3\linewidth]{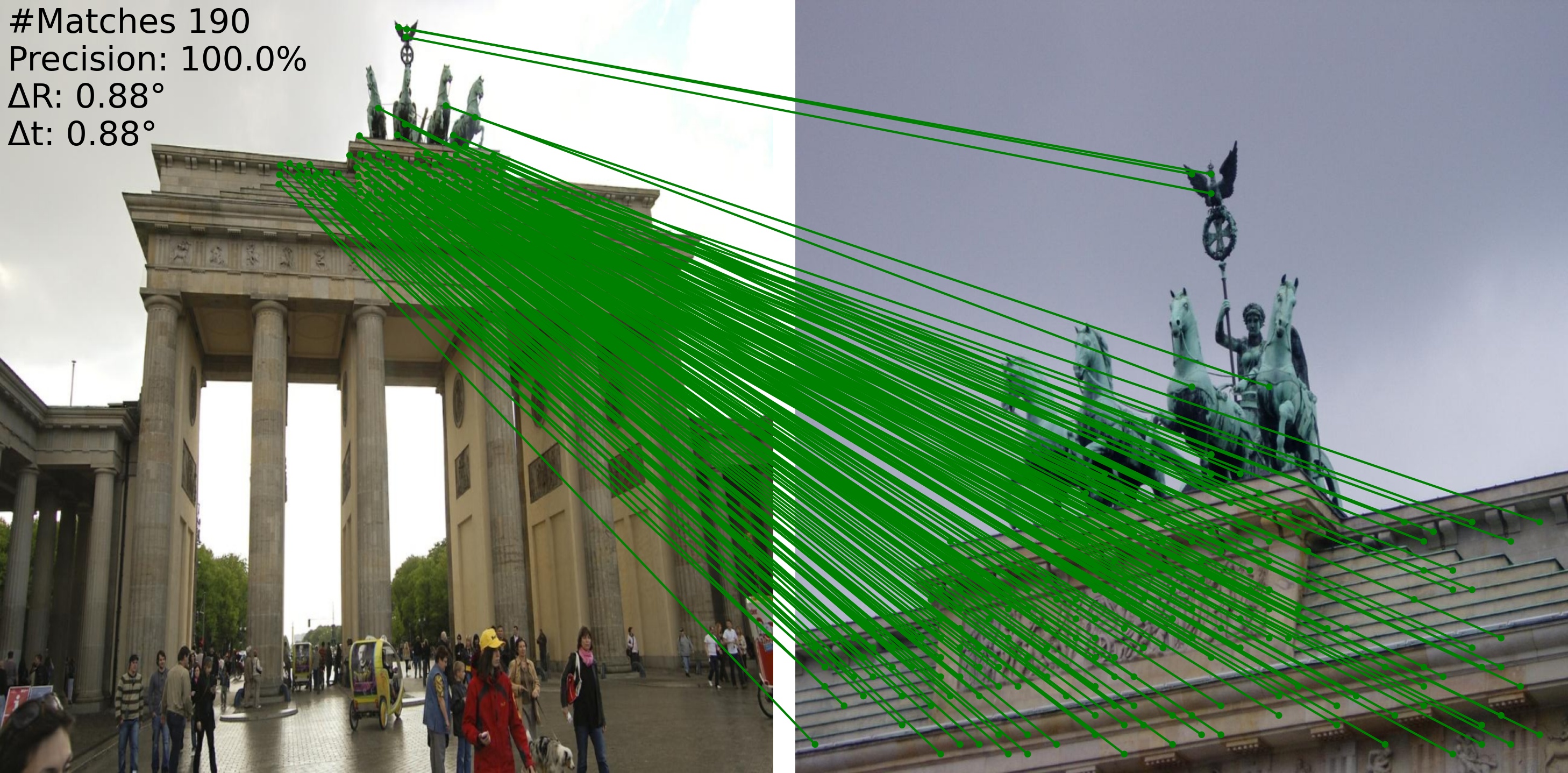} 
				& \includegraphics[width=0.3\linewidth]{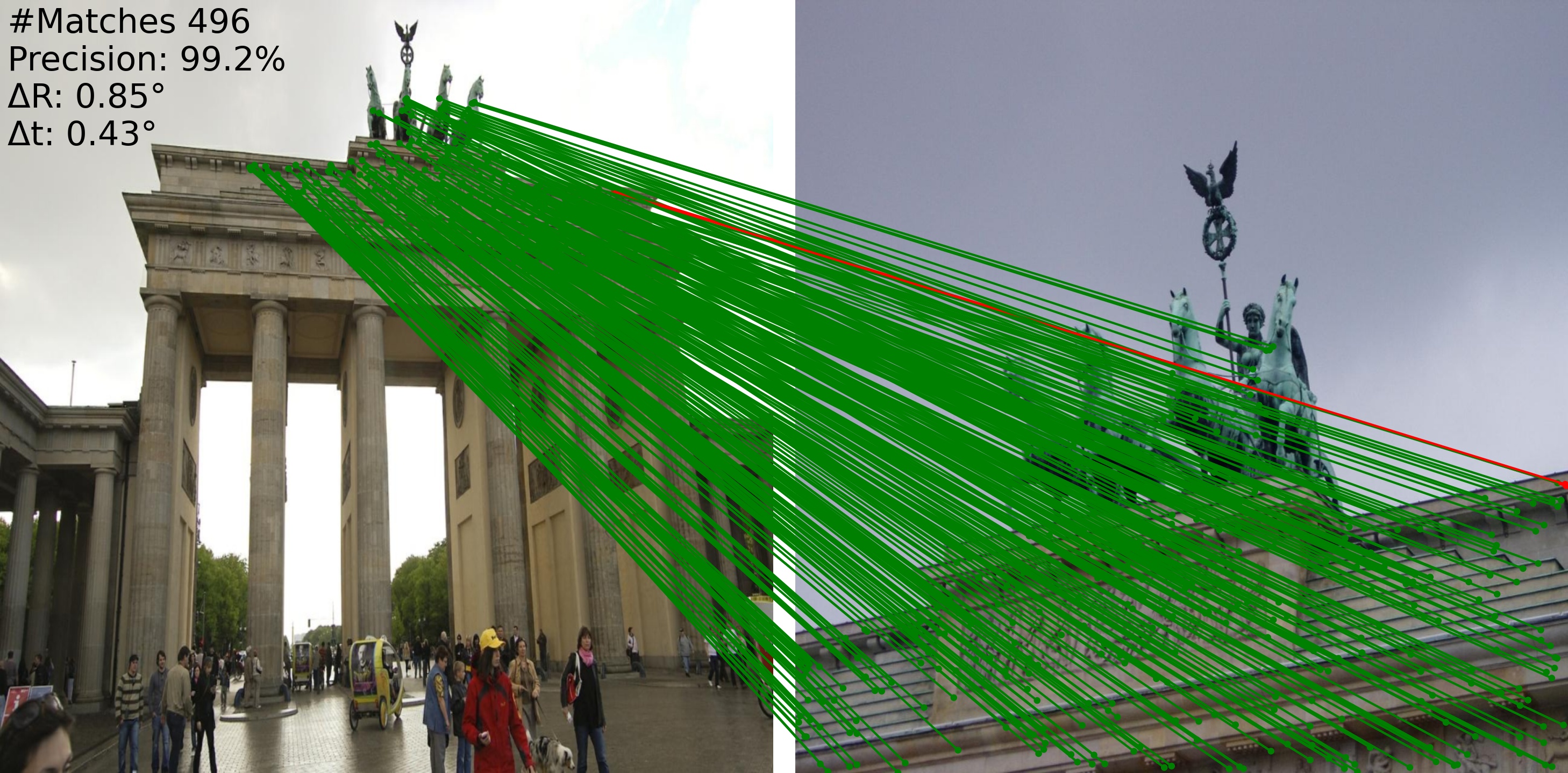} \\
				
				& \includegraphics[width=0.3\linewidth]{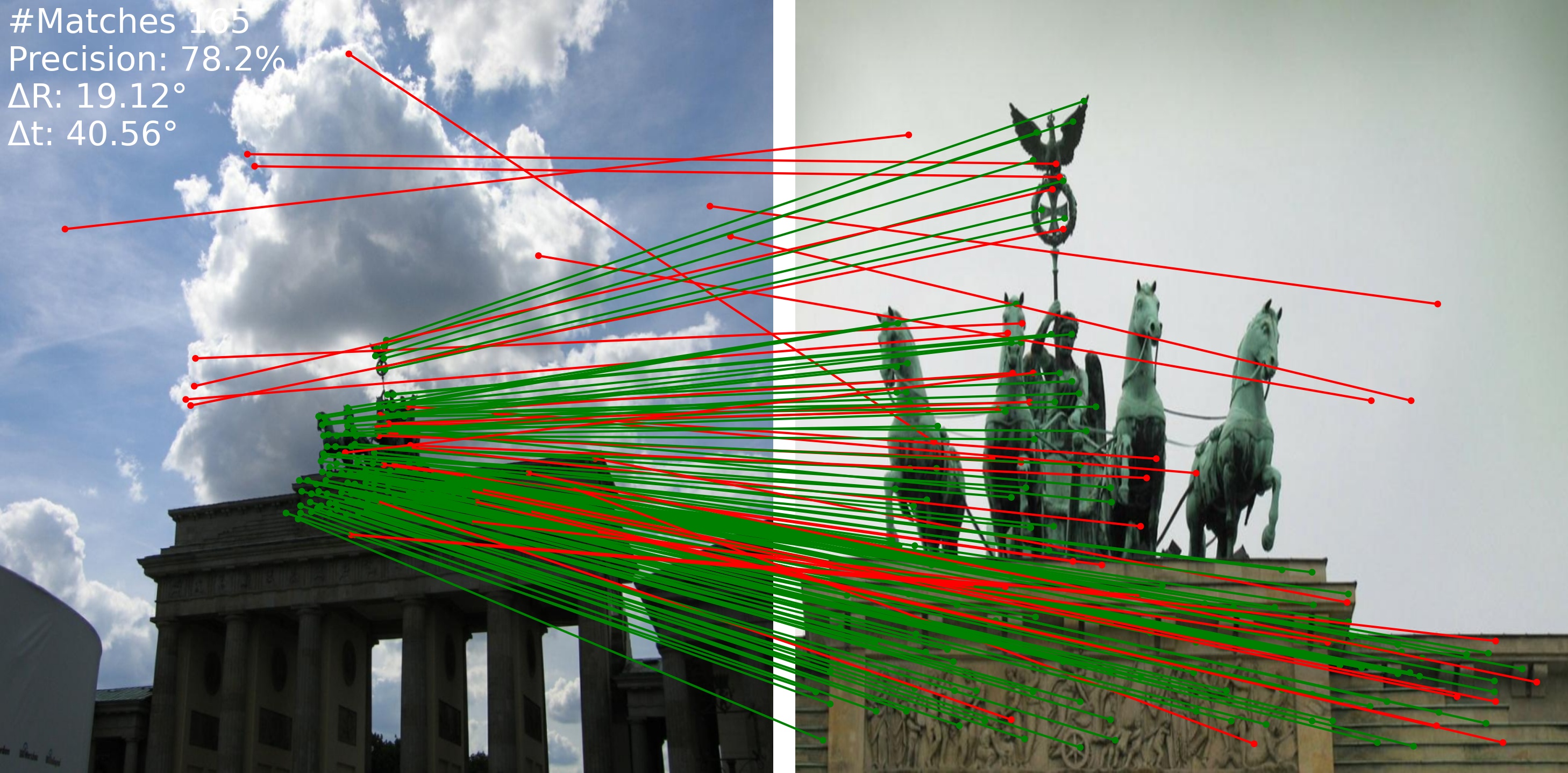}
				& \includegraphics[width=0.3\linewidth]{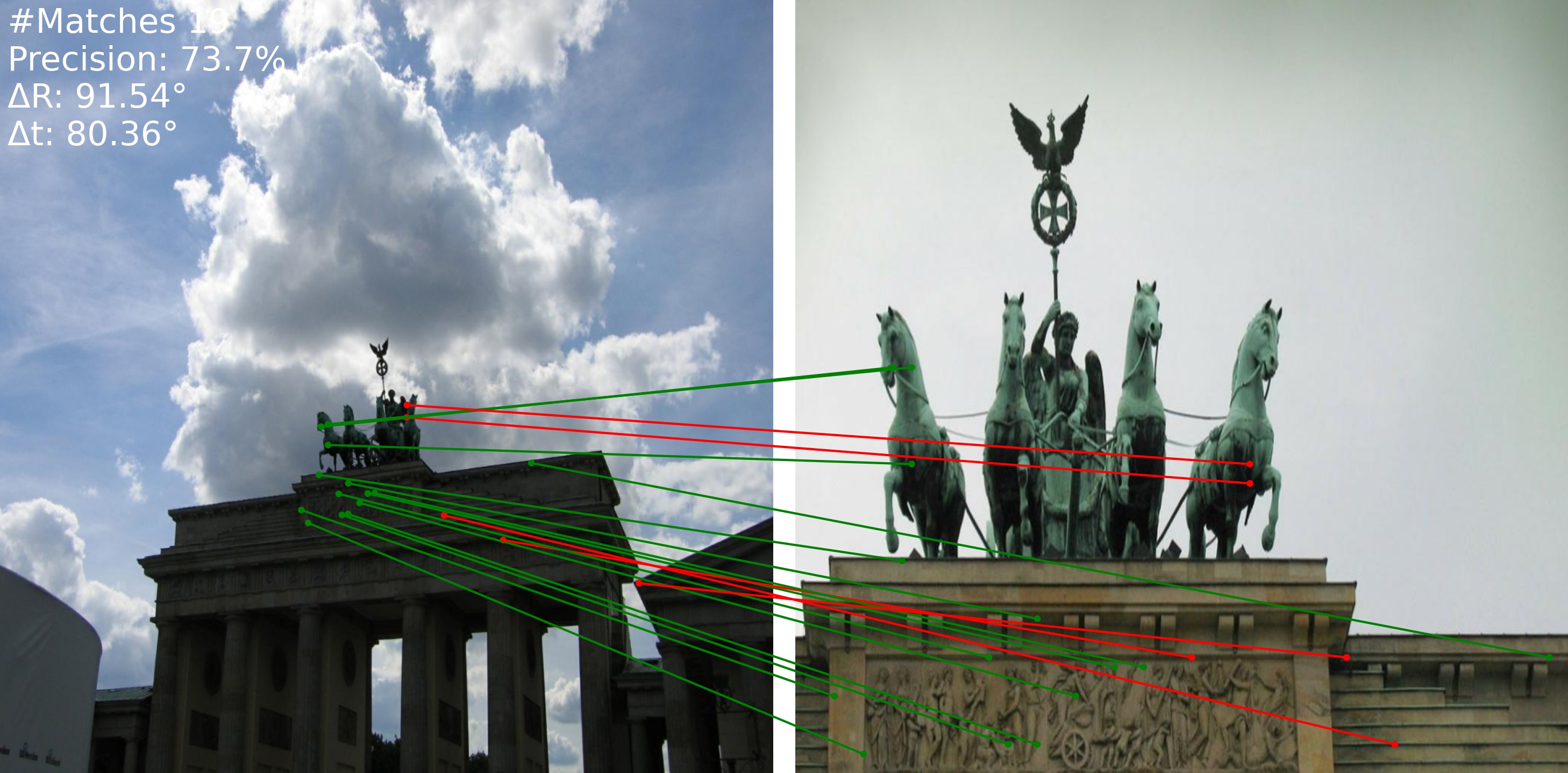} 
				& \includegraphics[width=0.3\linewidth]{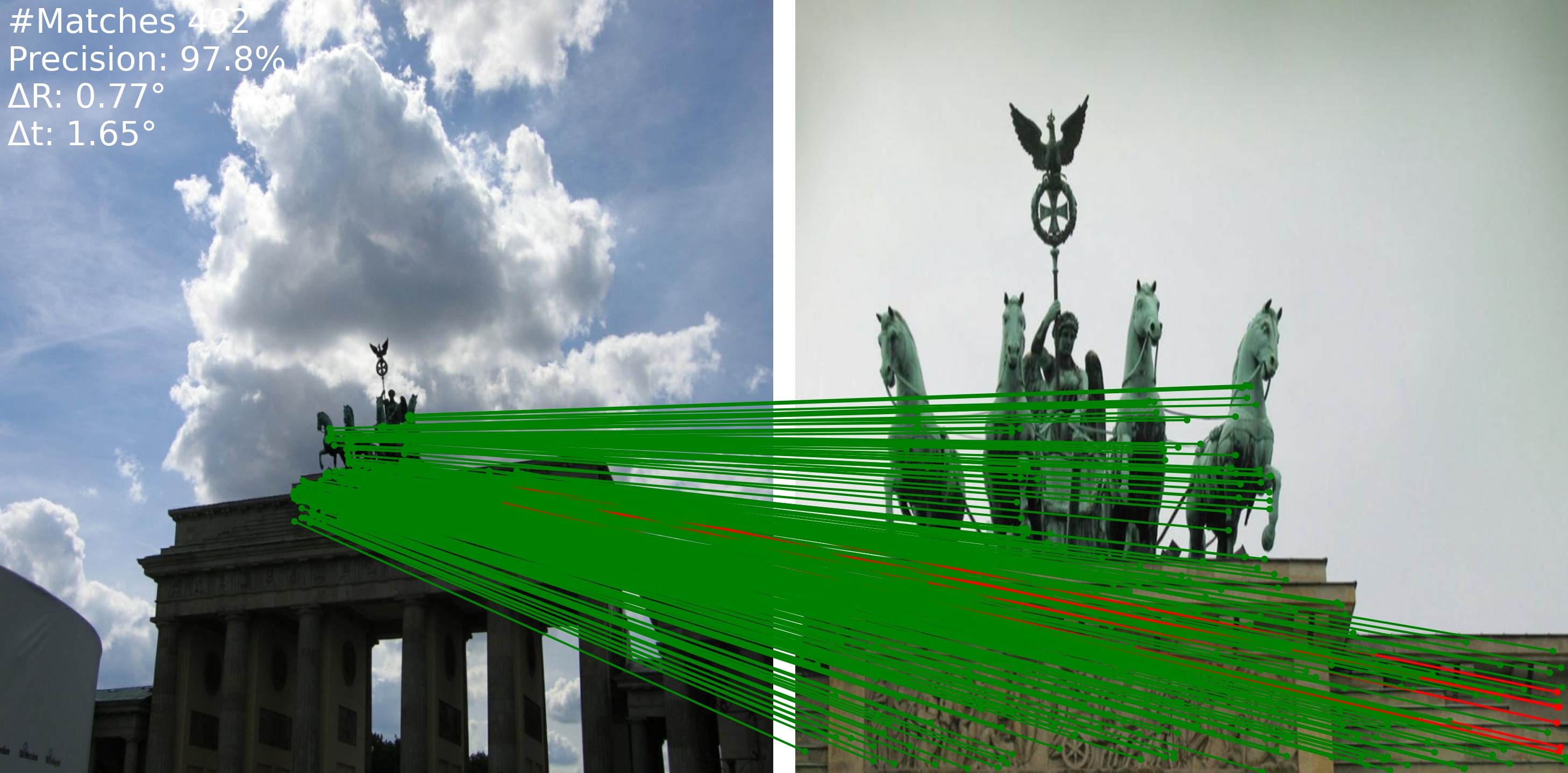} \\
				
				{\multirow{2}{*}{\rotatebox[origin=c]{90}{ScanNet}}} &
				\includegraphics[width=0.3\linewidth]{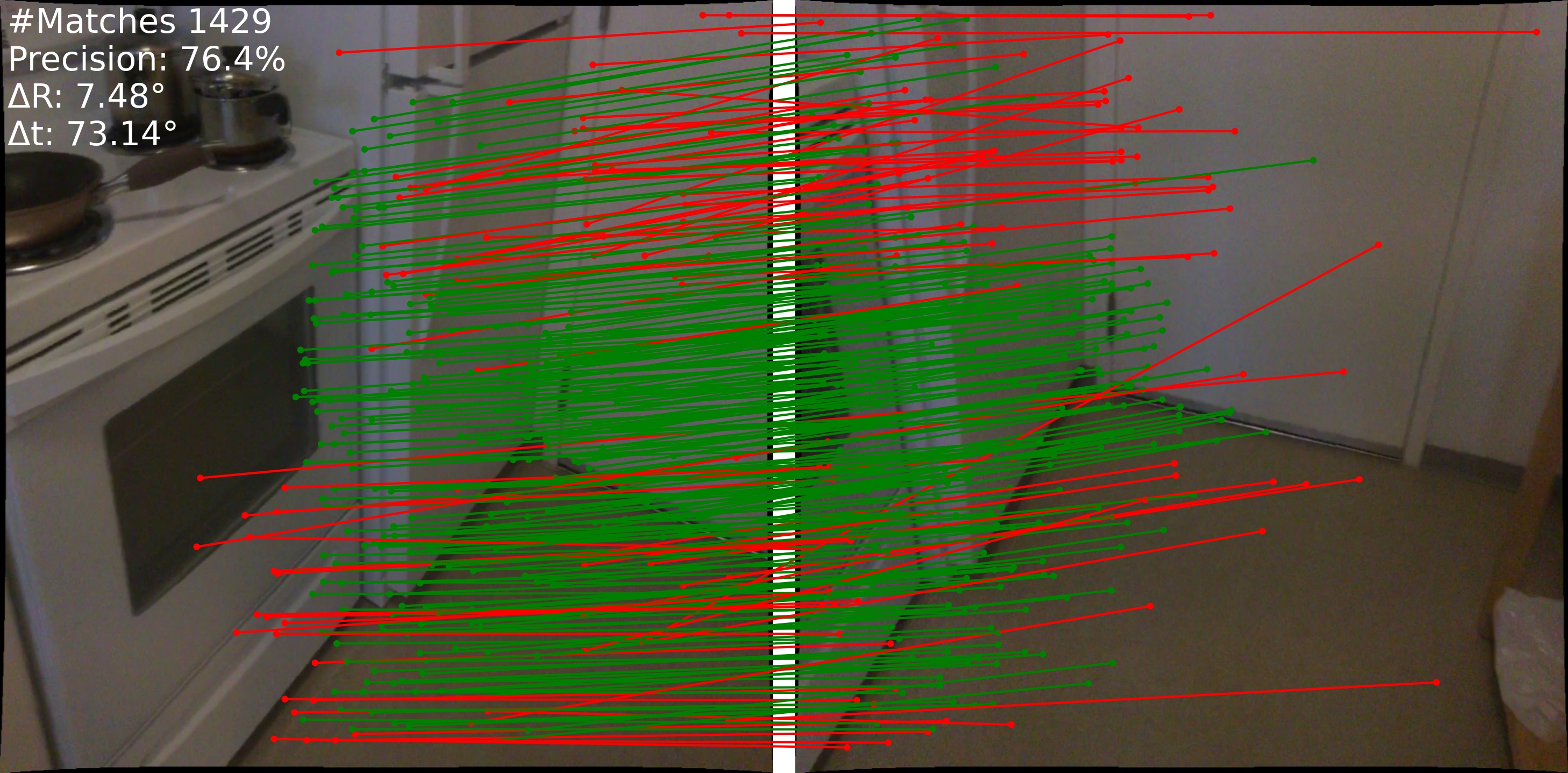}
				& \includegraphics[width=0.3\linewidth]{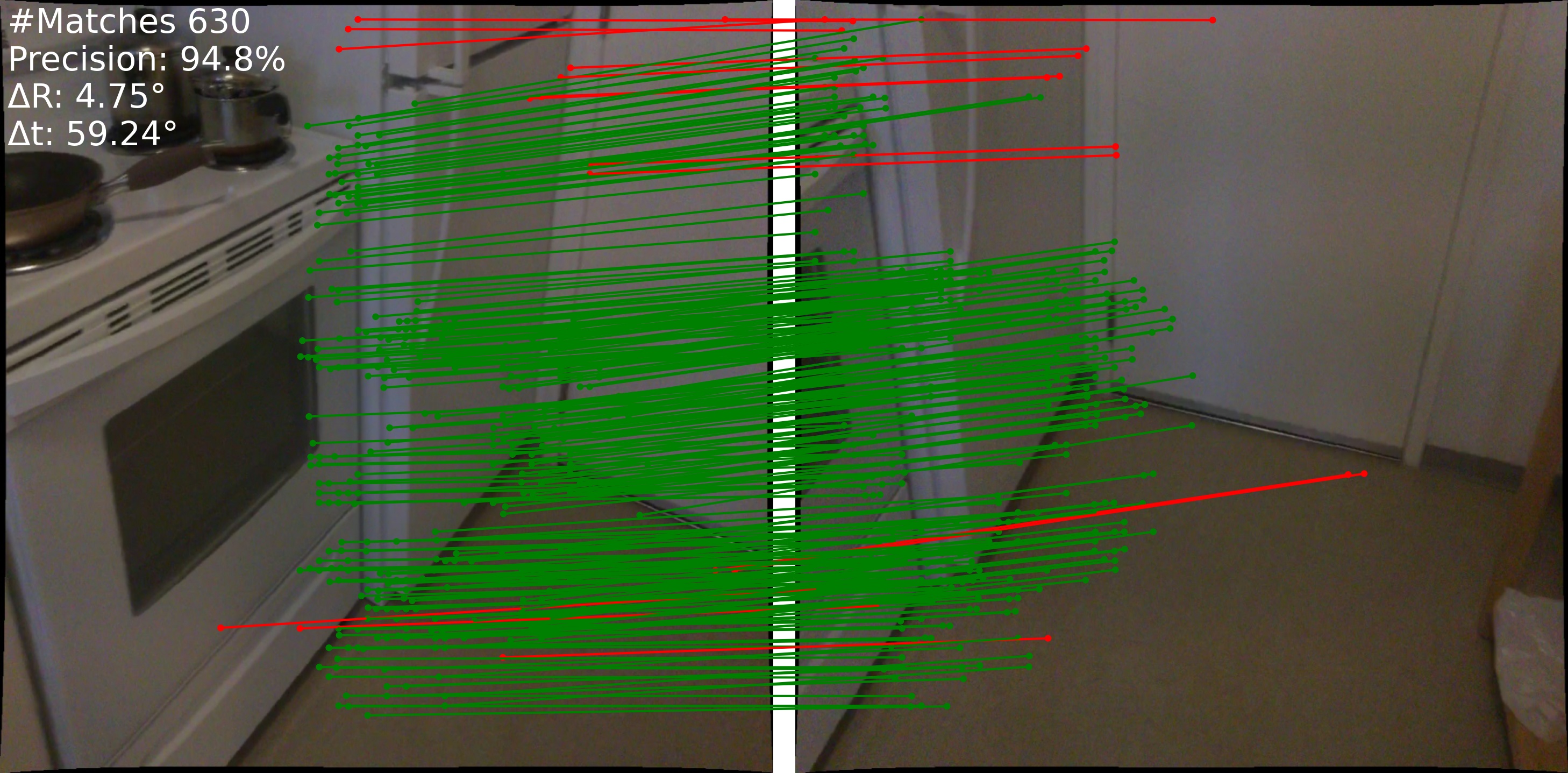} 
				& \includegraphics[width=0.3\linewidth]{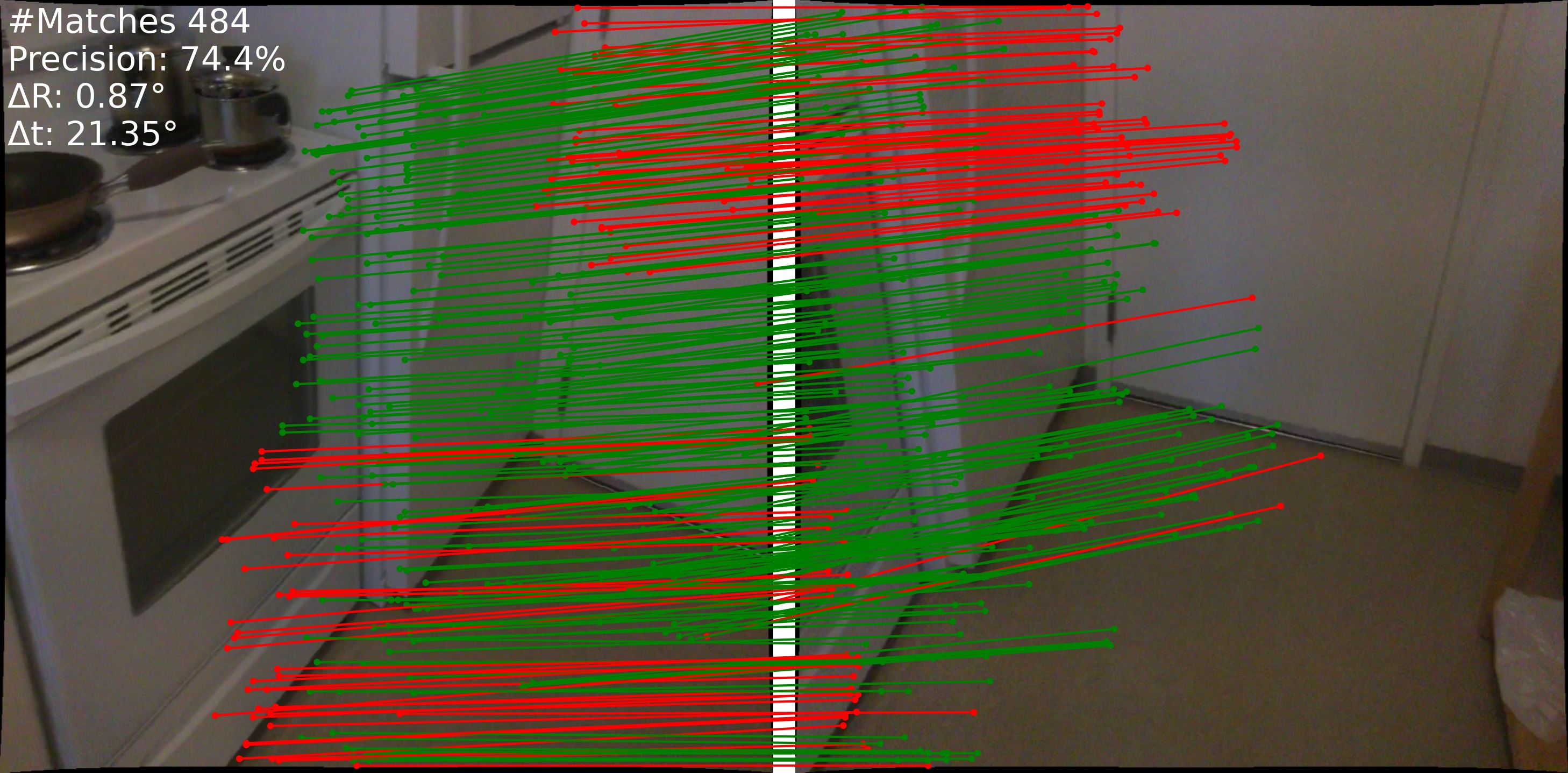} \\

				& \includegraphics[width=0.3\linewidth]{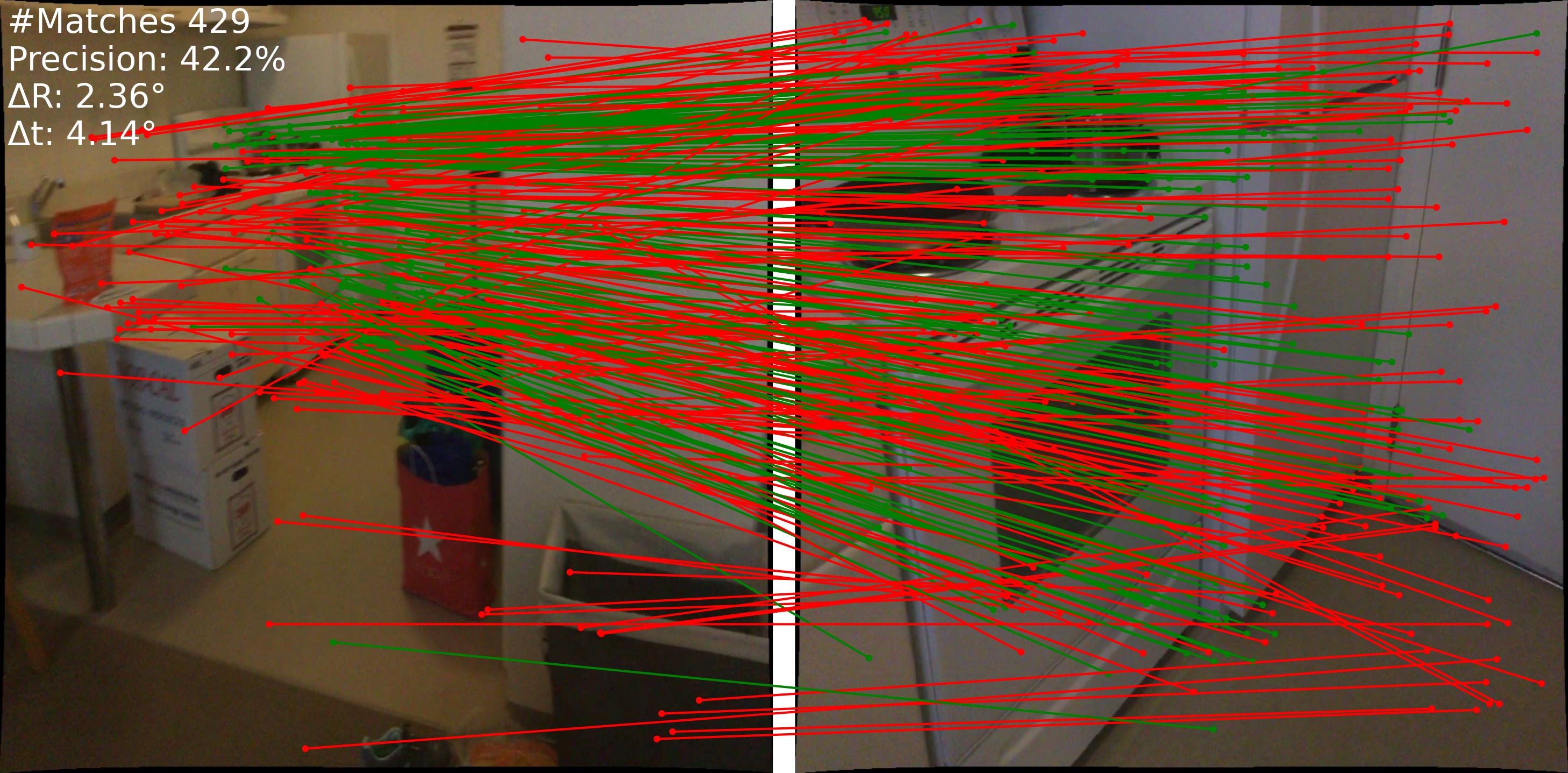}
				& \includegraphics[width=0.3\linewidth]{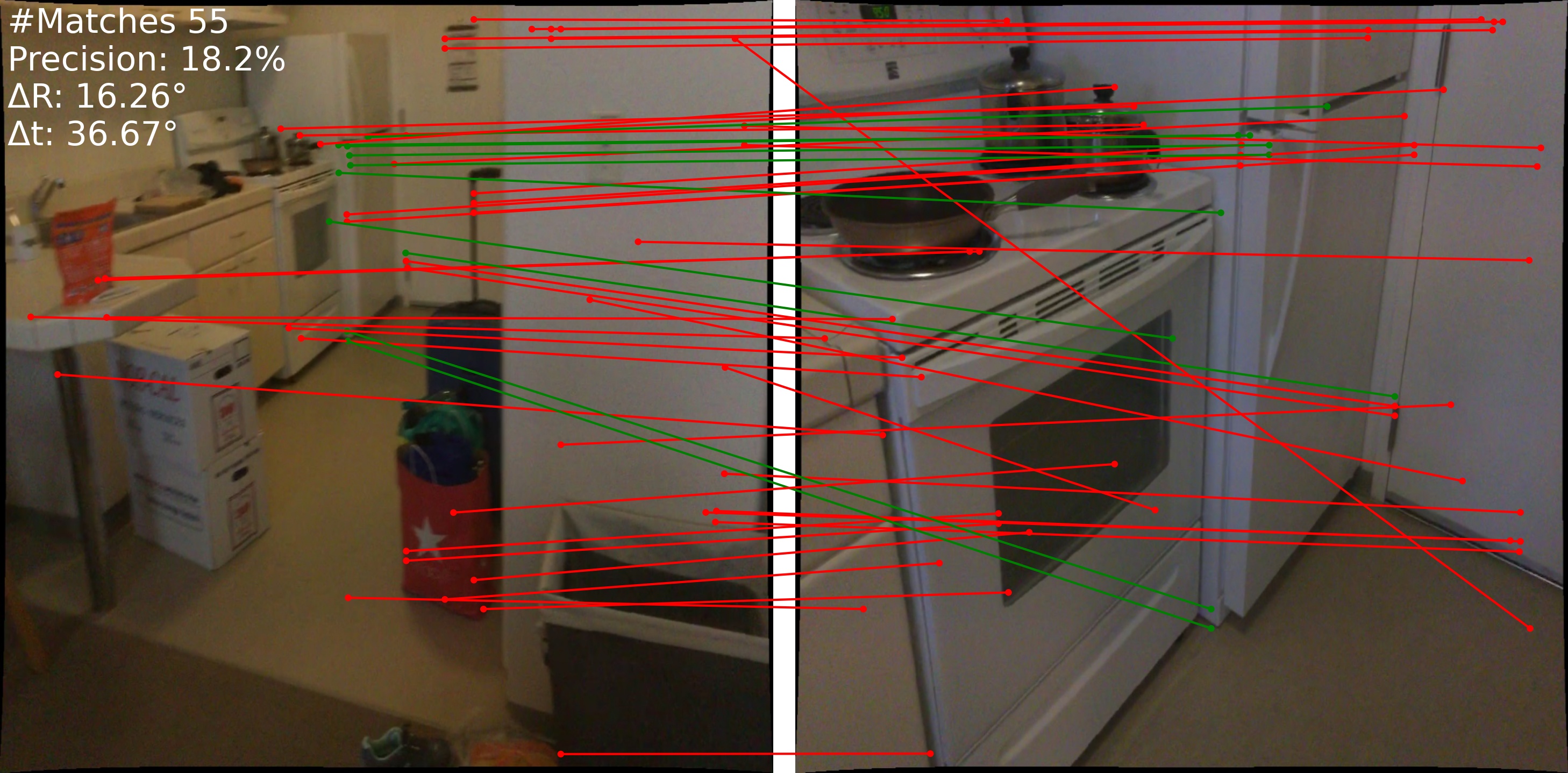} 
				& \includegraphics[width=0.3\linewidth]{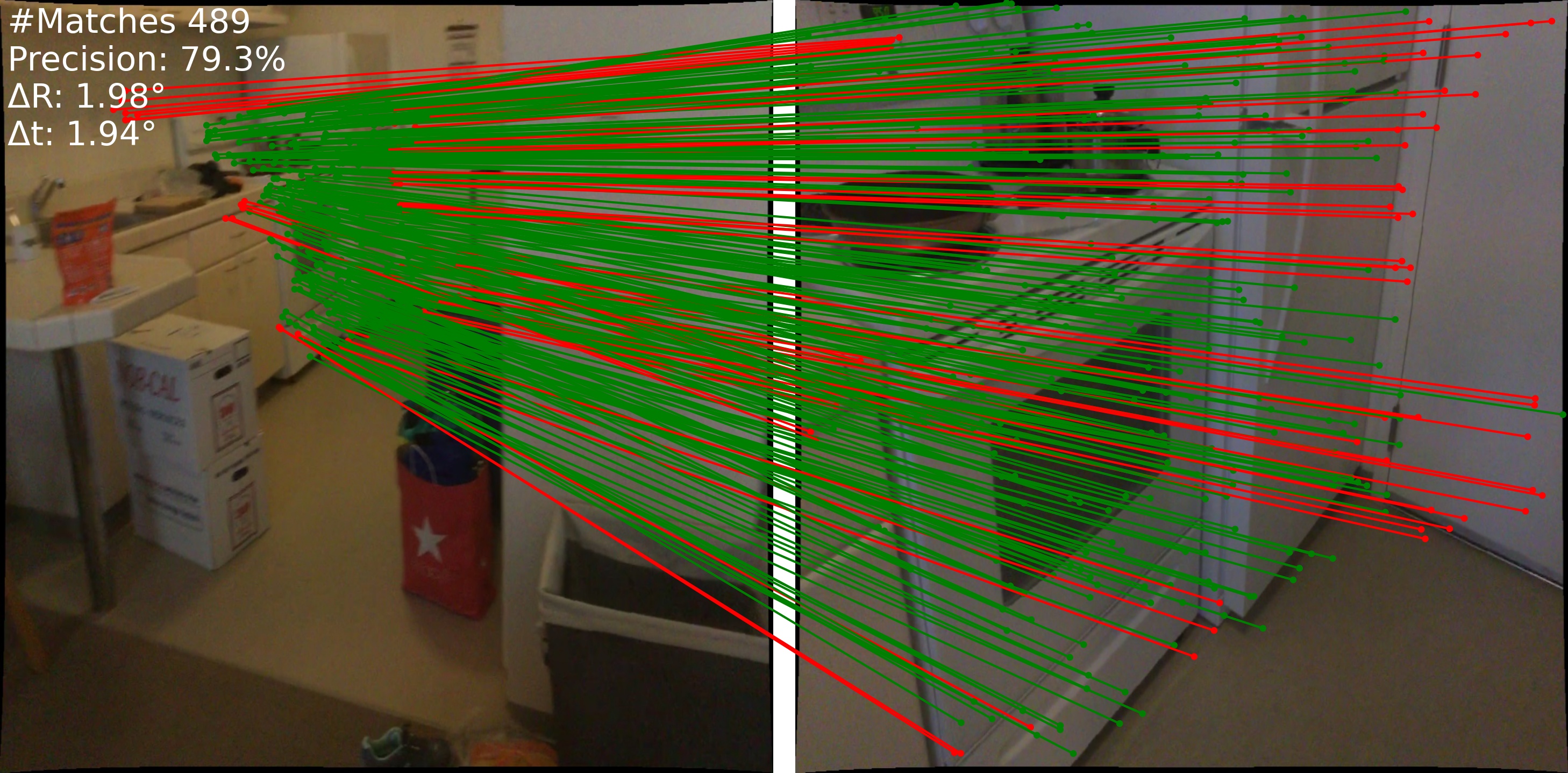} \\	

				{\multirow{2}{*}{\rotatebox[origin=c]{90}{TartanAir}}} &
\includegraphics[width=0.3\linewidth]{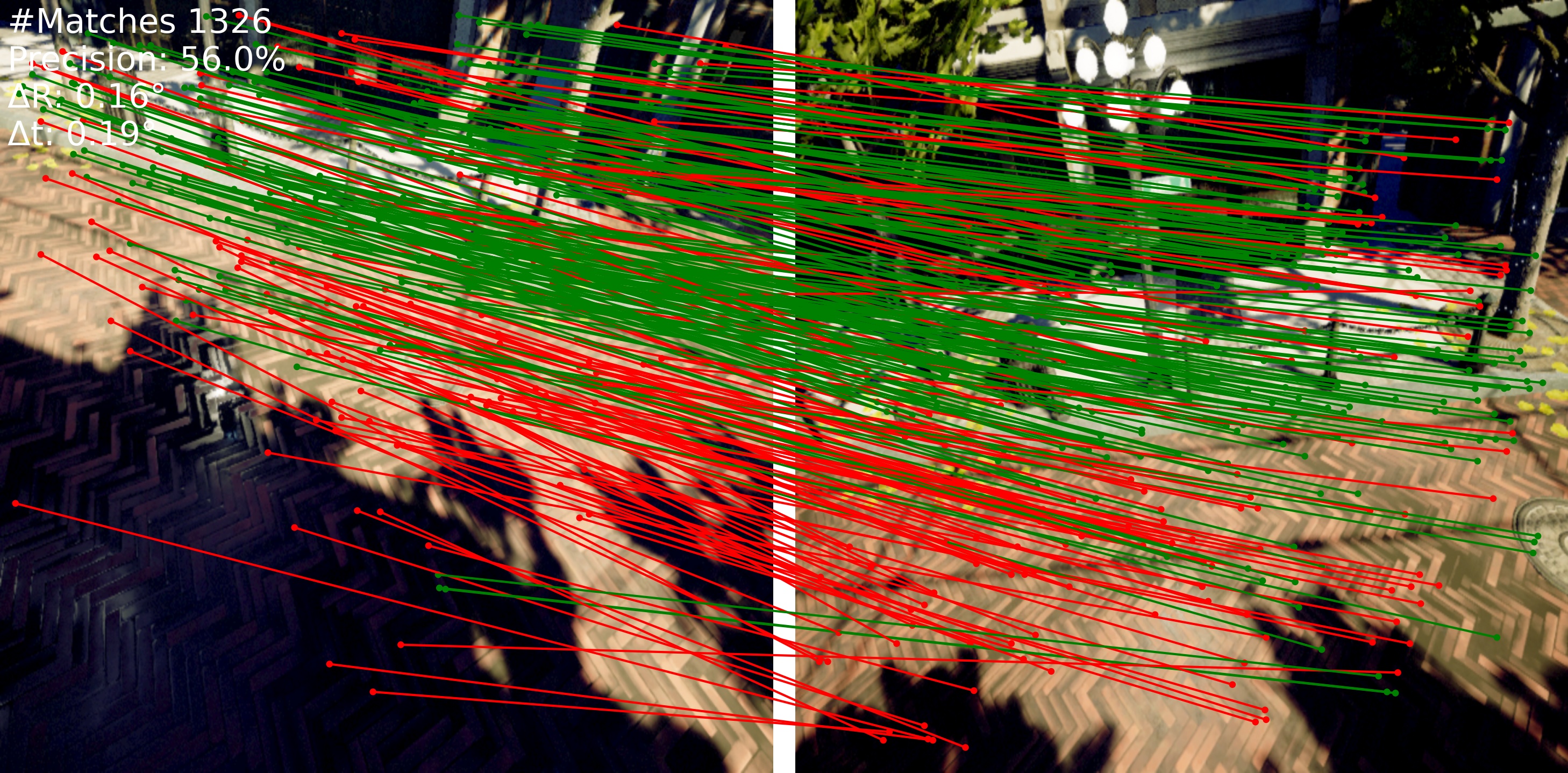}
& \includegraphics[width=0.3\linewidth]{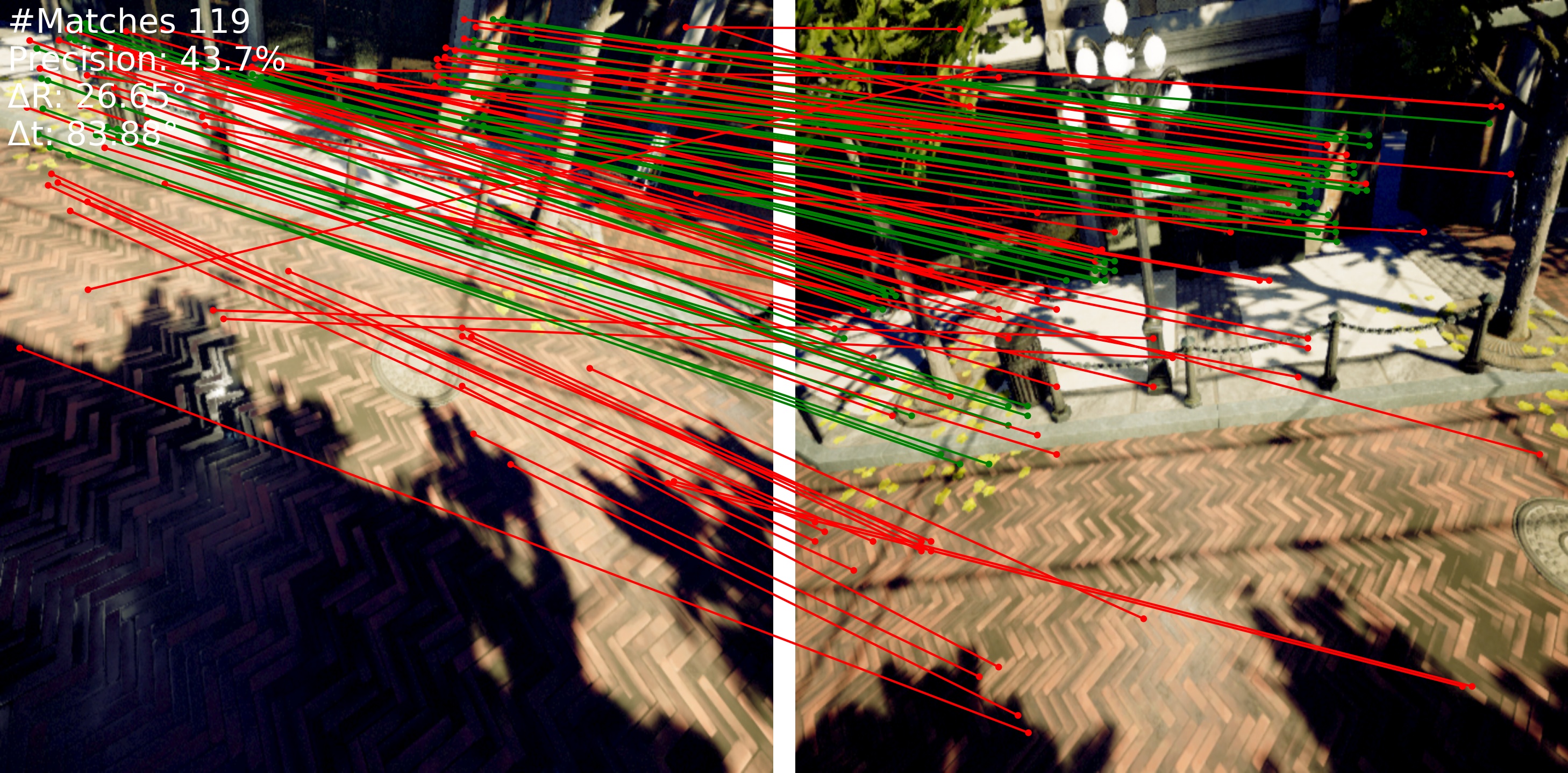} 
& \includegraphics[width=0.3\linewidth]{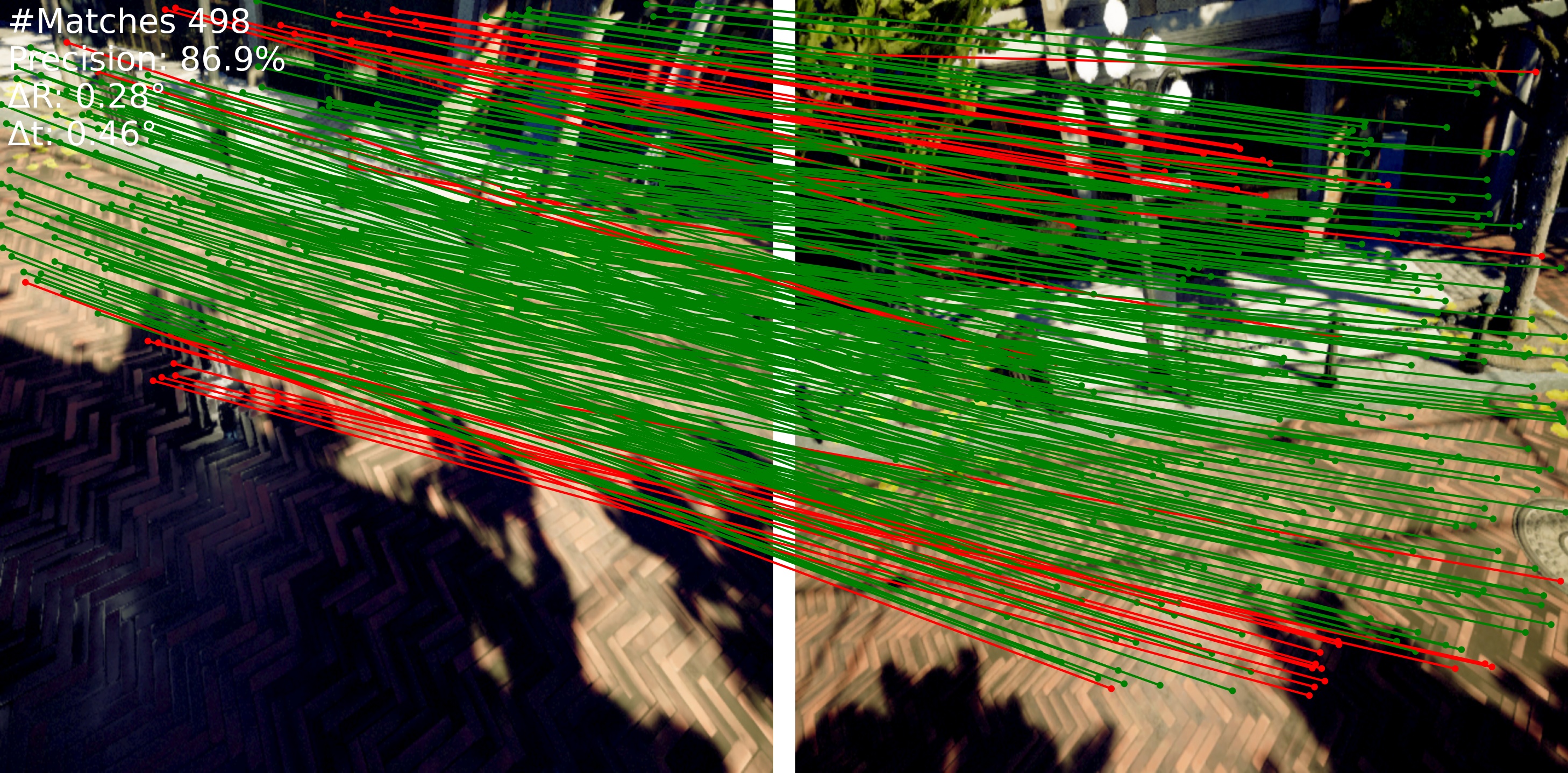} \\

& \includegraphics[width=0.3\linewidth]{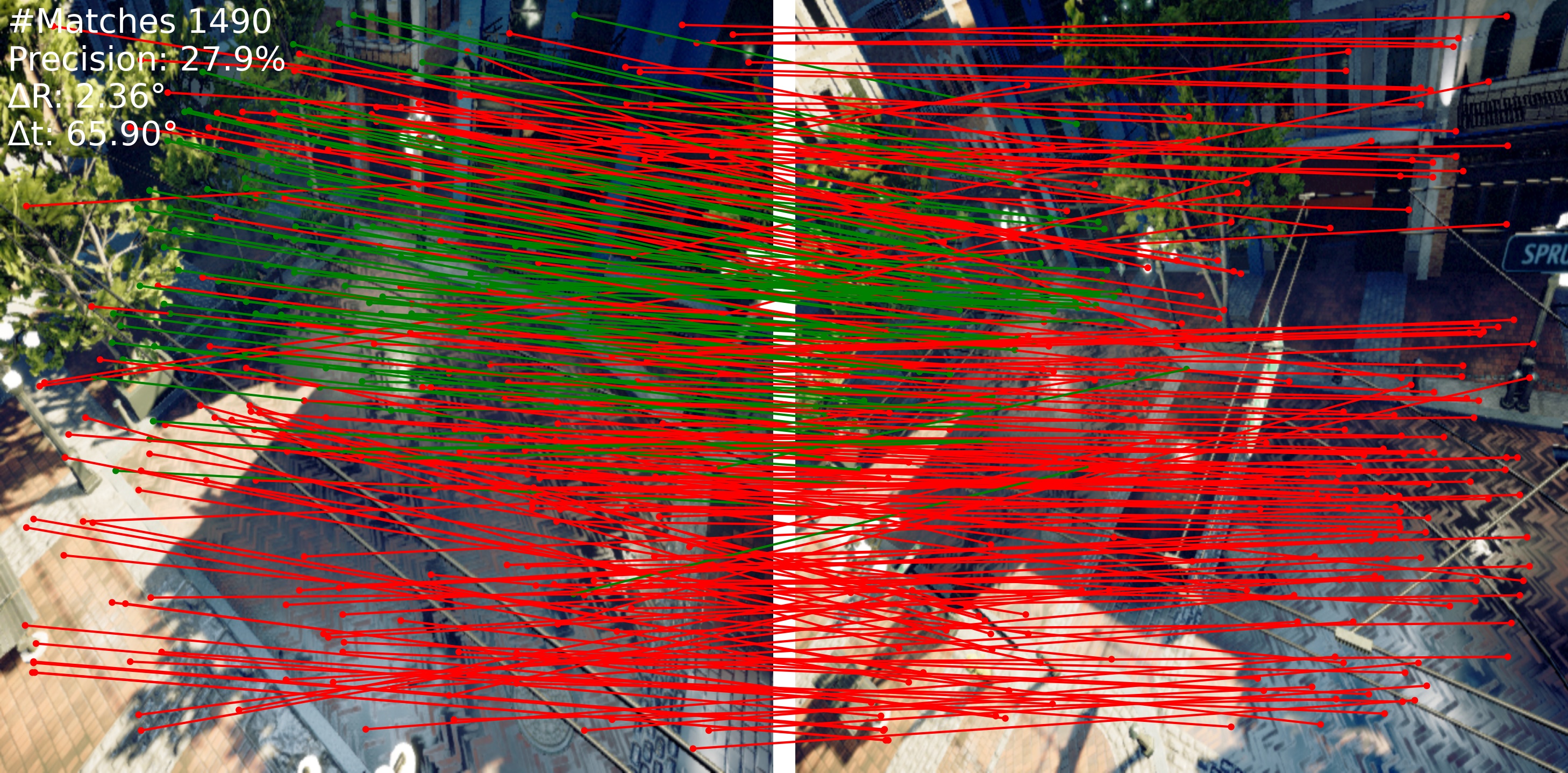}
& \includegraphics[width=0.3\linewidth]{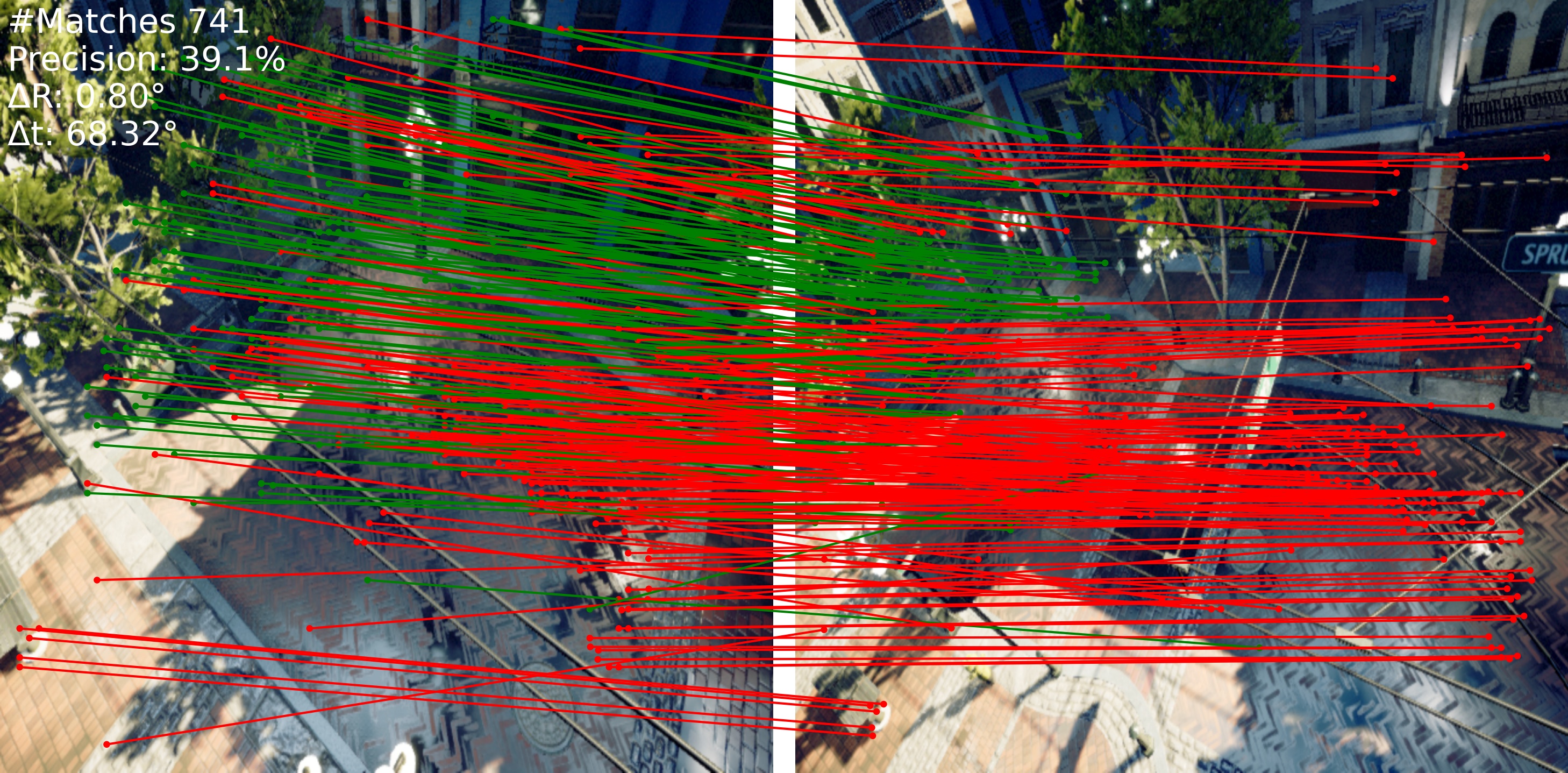} 
& \includegraphics[width=0.3\linewidth]{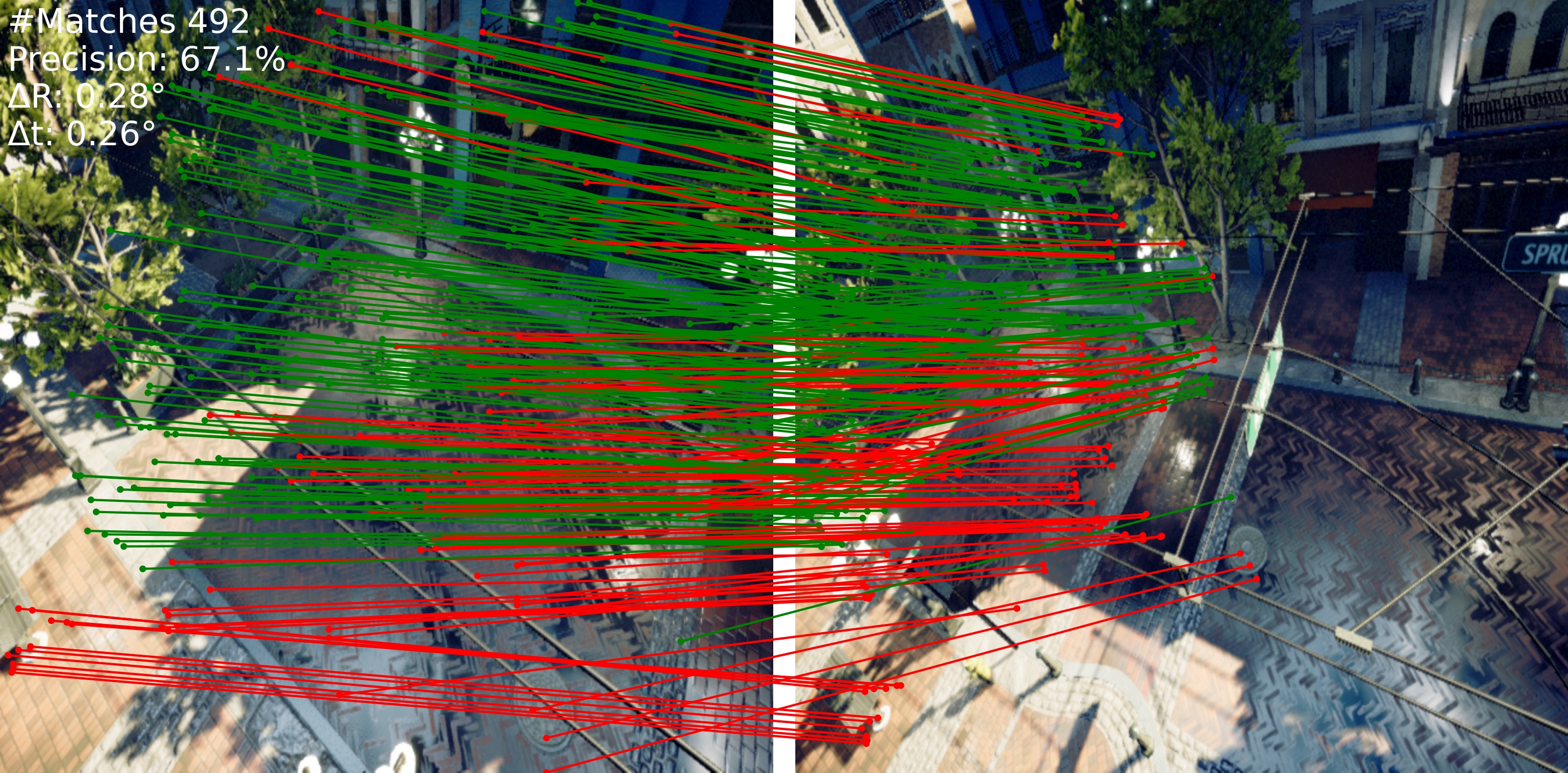} \\
			
				\bottomrule
			\end{tabular}
		\end{small}
	\end{center}
\caption{
	Qualitative comparison under wide-baseline geometric scenes and challenging robotic perception scenarios. Red indicates an epipolar error greater than $5\times10^{-4}$ in normalized image coordinates. CoReMatch produces more consistent correspondences across challenging appearance and scene variations than ELoFTR and EDM.
}
\label{tab_a_qualitative}
\end{figure}

\begin{figure}[t]
	\begin{center}
			\begin{small}
					\begin{tabular}{ccc}
							\toprule
							ELoFTR & EDM & Ours\\
							\midrule
							
							\includegraphics[width=0.3\linewidth]{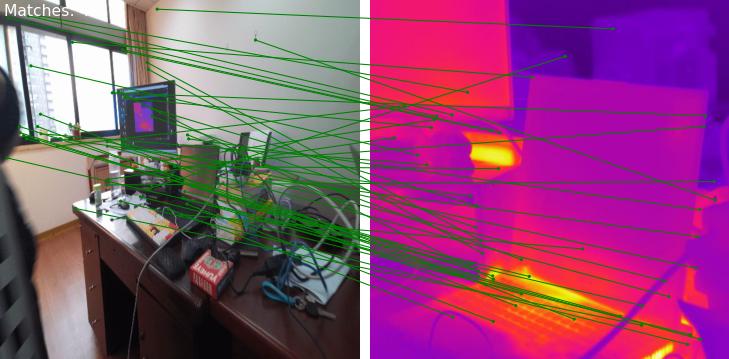}
							& 
							\includegraphics[width=0.3\linewidth]{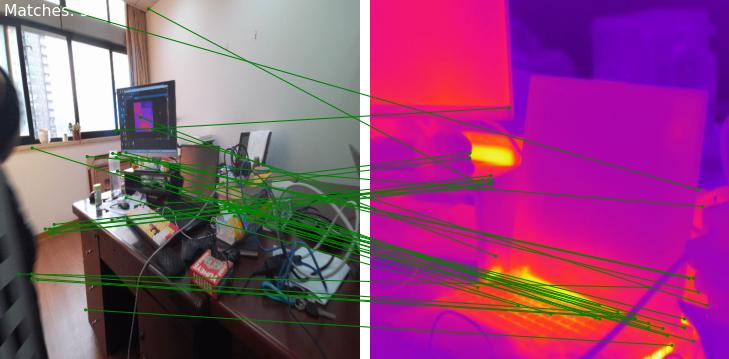} 
							& 
							\includegraphics[width=0.3\linewidth]
							{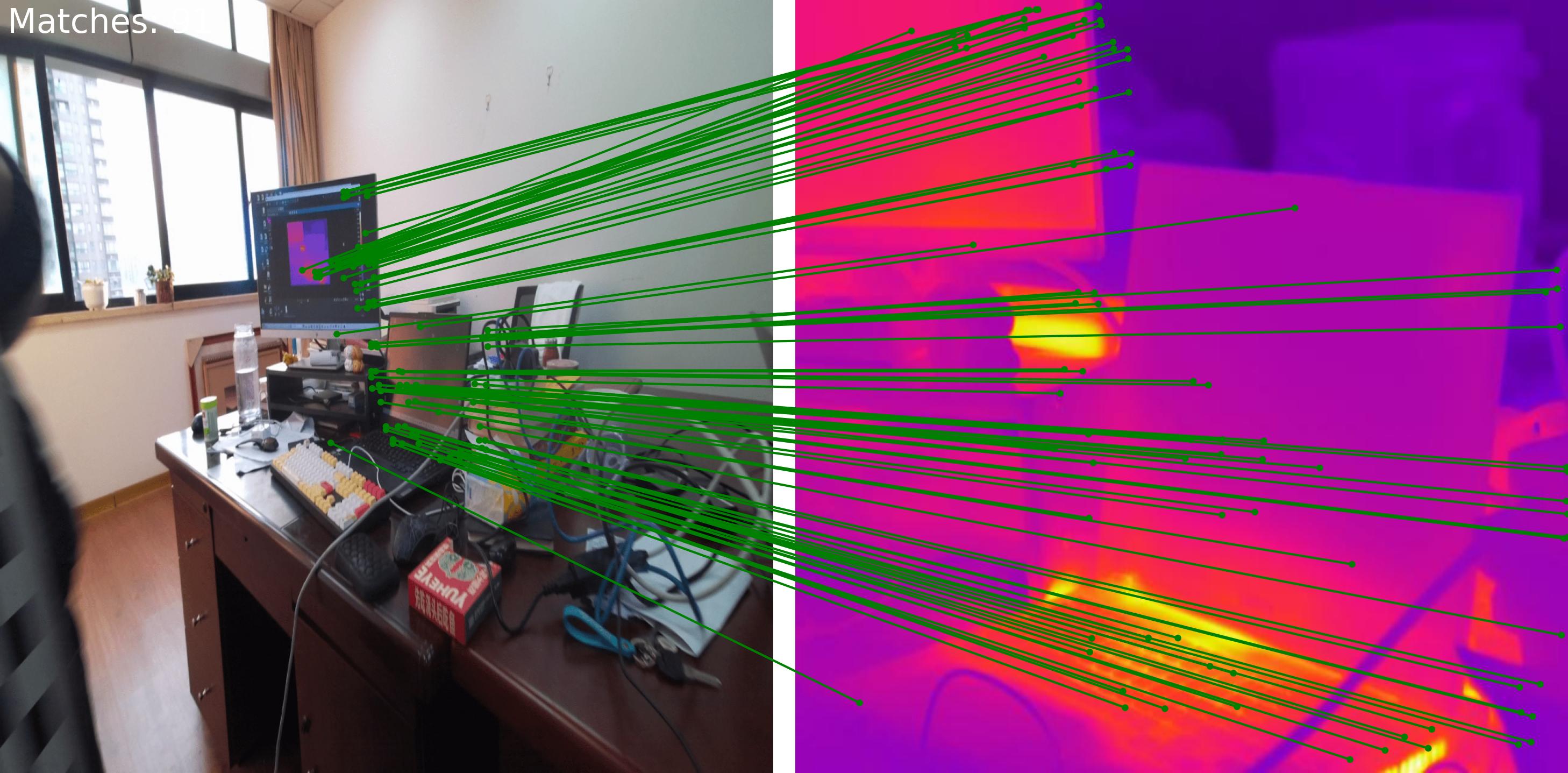} \\
							
							\includegraphics[width=0.3\linewidth]{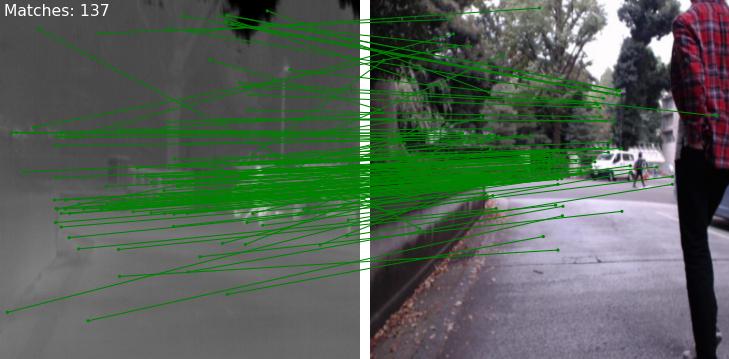}
							& 
							\includegraphics[width=0.3\linewidth]{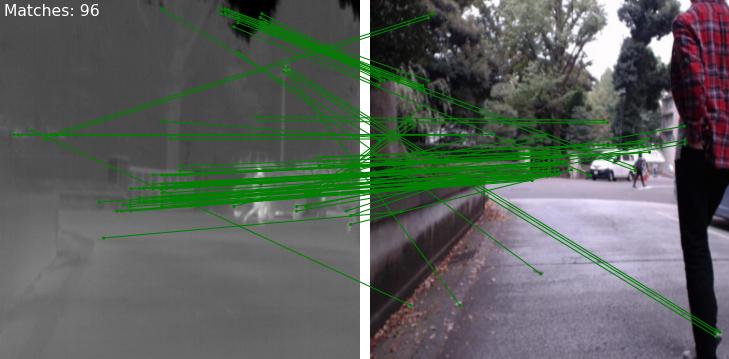} 
							& 
							\includegraphics[width=0.3\linewidth]
							{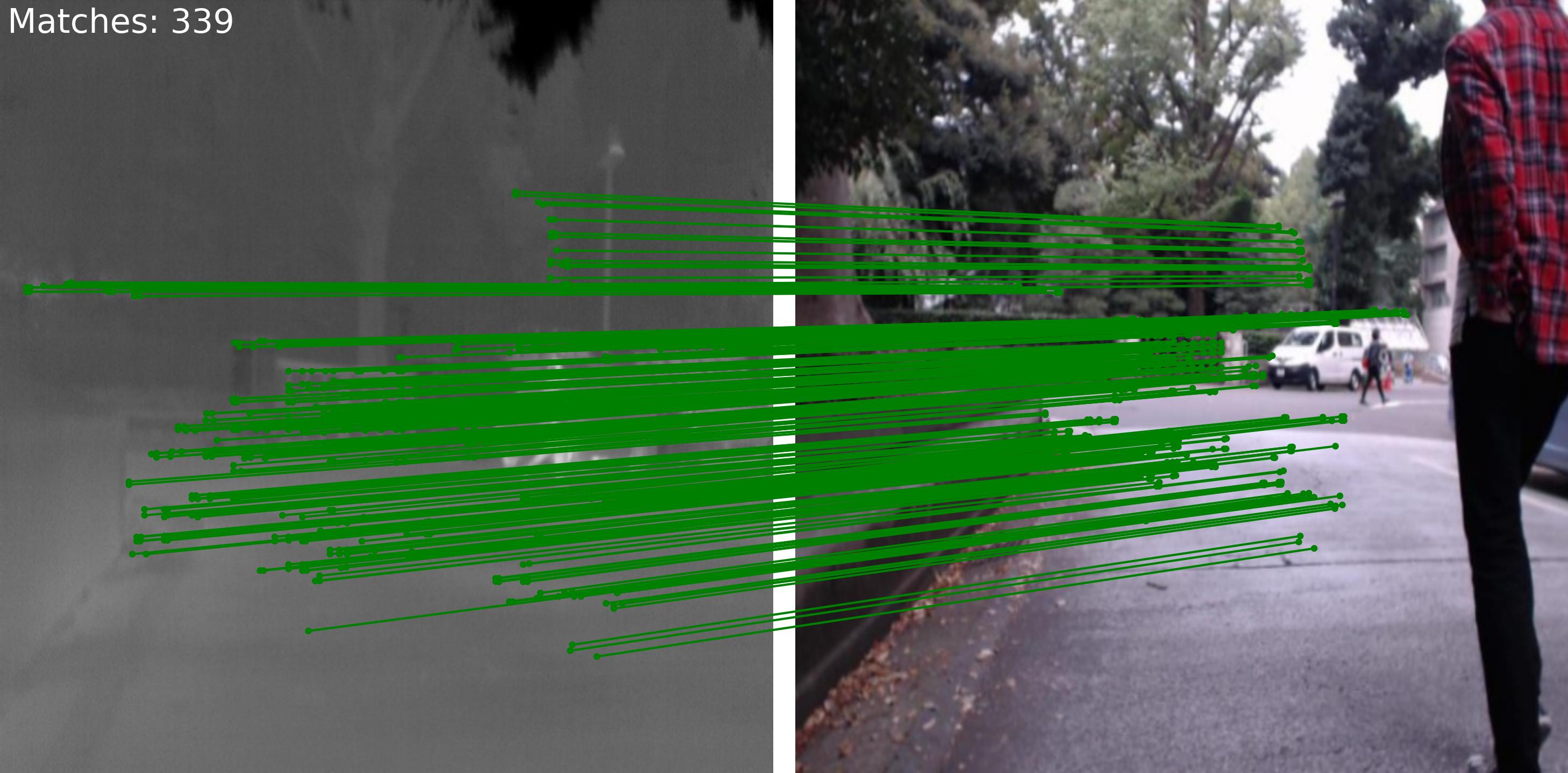} \\
							
							\includegraphics[width=0.3\linewidth]{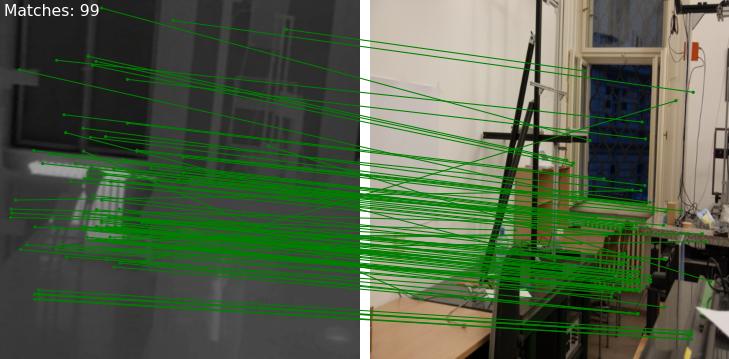}
							& 
							\includegraphics[width=0.3\linewidth]{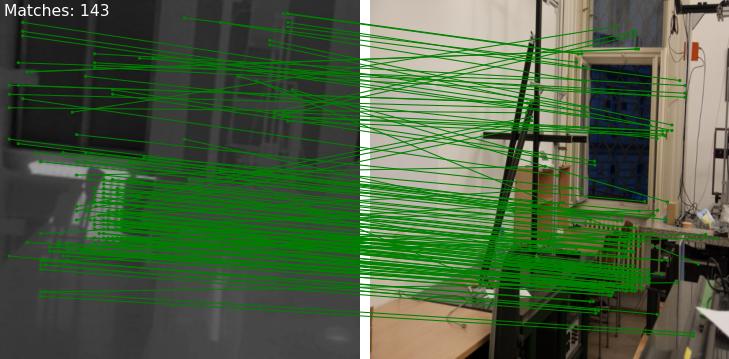} 
							& 
							\includegraphics[width=0.3\linewidth]
							{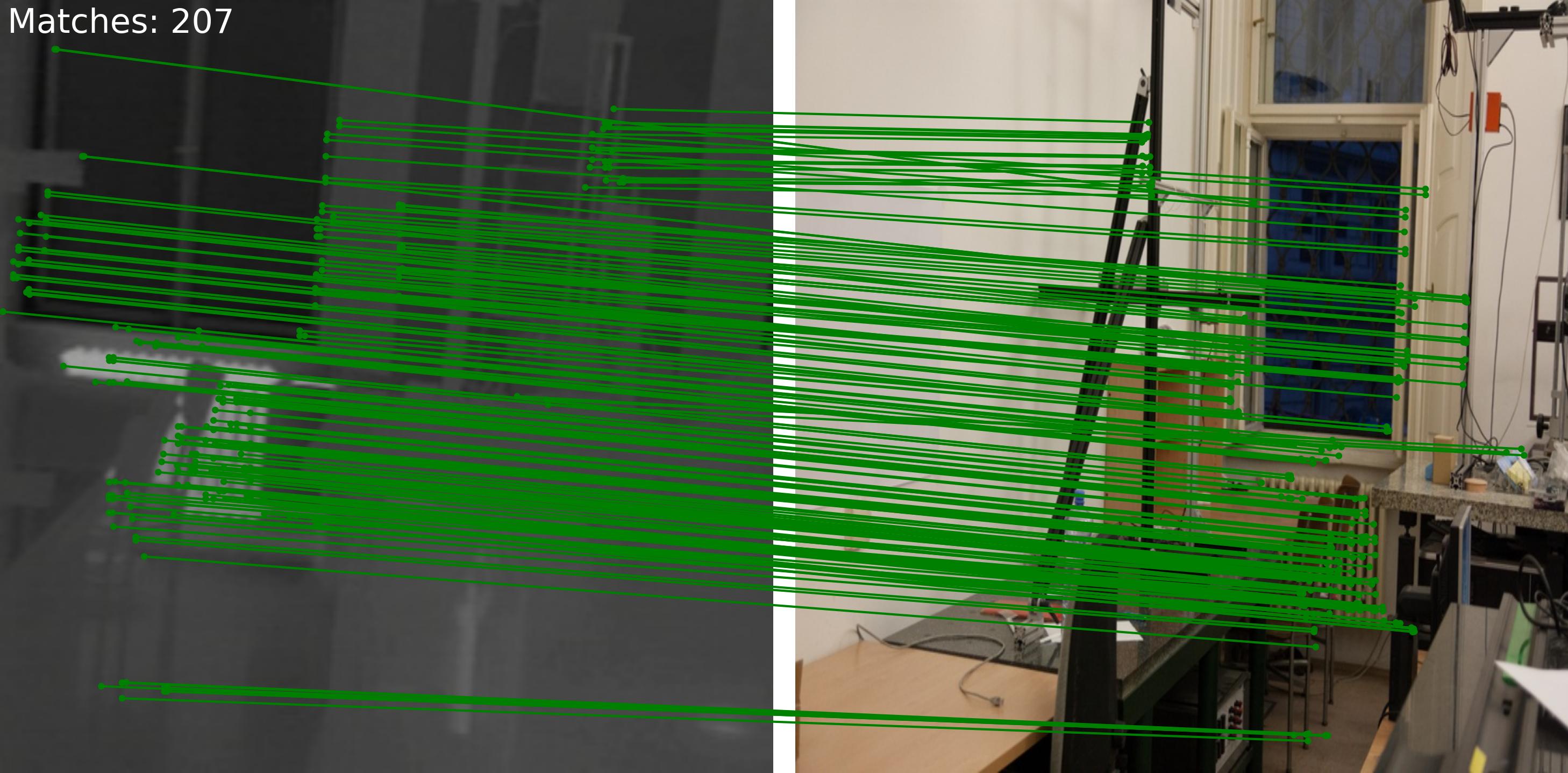} \\
							
							\bottomrule
						\end{tabular}
				\end{small}
		\end{center}
	\caption{
	\textbf{Qualitative comparison on visible--thermal cross-modal matching}.
	CoReMatch demonstrates stronger robustness and produces more reliable correspondences across severe modality differences.
	In contrast, ELoFTR and EDM generate substantially more incorrect matches, highlighting the effectiveness of recurrent coordinate refinement and robust feature representations for cross-modal matching.
}
	\label{tab_b_qualitative}
\end{figure}

We qualitatively compare Efficient LoFTR (ELoFTR), EDM, and CoReMatch-M under challenging robotic perception scenarios, including motion blur, large viewpoint changes, repetitive structures, low-texture regions, and visible--thermal cross-modal matching. As shown in Fig.~\ref{tab_a_qualitative} and Fig.~\ref{tab_b_qualitative}, CoReMatch produces more consistent and reliable correspondences across wide-baseline geometric scenes and substantial appearance and modality variations. On MegaDepth and ScanNet, CoReMatch maintains reliable correspondences under changes in illumination and scene content, whereas ELoFTR and EDM can produce accurate matches in one view but fail under such variations. On the challenging TartanAir scenarios, CoReMatch consistently produces more reliable correspondences than both baselines. These results demonstrate the robustness of CoReMatch-M to challenging appearance variations and cross-modal matching.

\begin{figure}[t]
	\begin{center}
		\footnotesize
		\begin{tabular}{lc}
						\raisebox{-0.5\height}{\rotatebox{90}{Indoor}} &
		\raisebox{-0.5\height}{\includegraphics[width=0.3\linewidth]
			{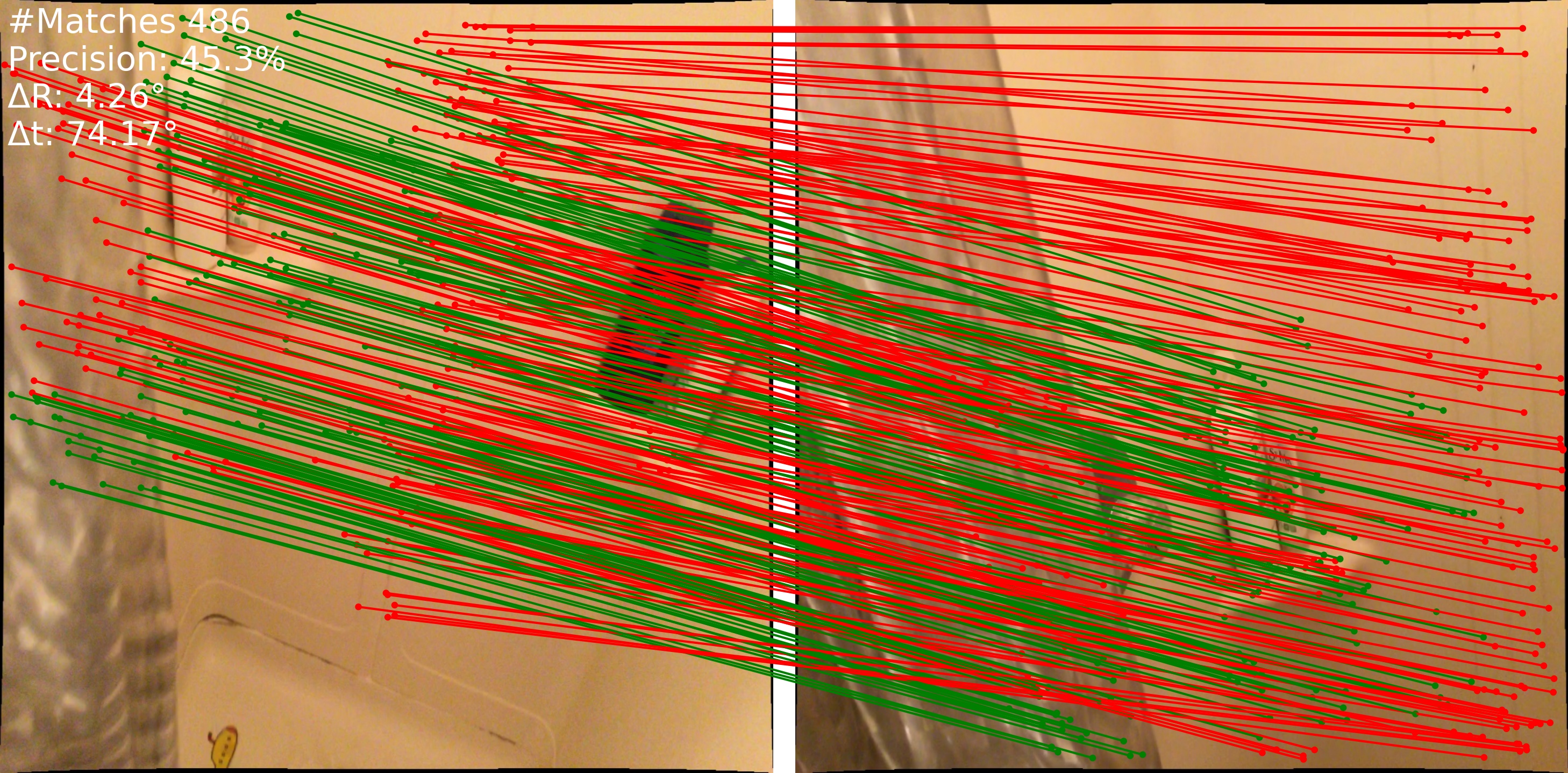}}
		\raisebox{-0.5\height}{\includegraphics[width=0.3\linewidth]{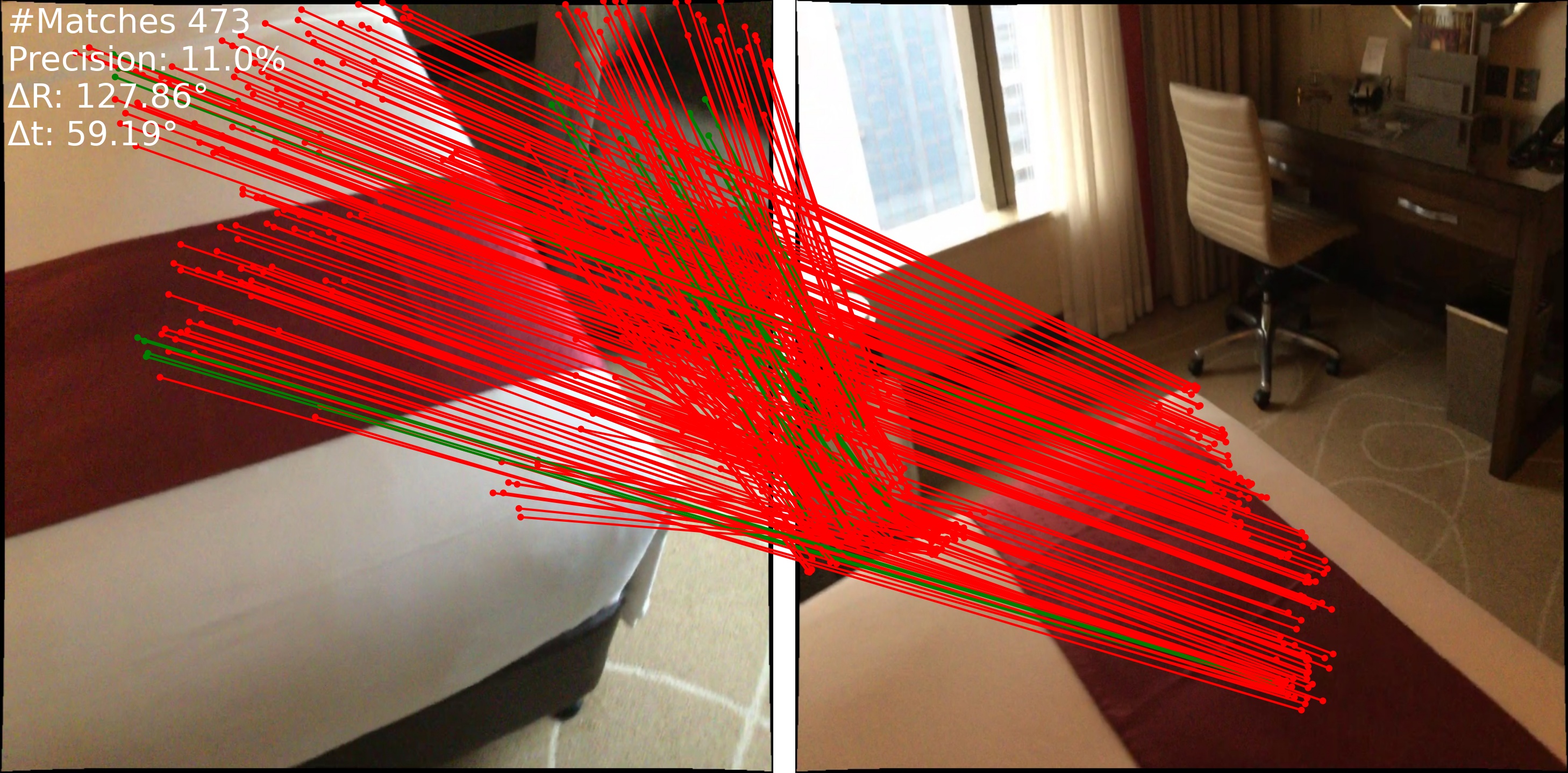}}
		\raisebox{-0.5\height}{\includegraphics[width=0.3\linewidth]{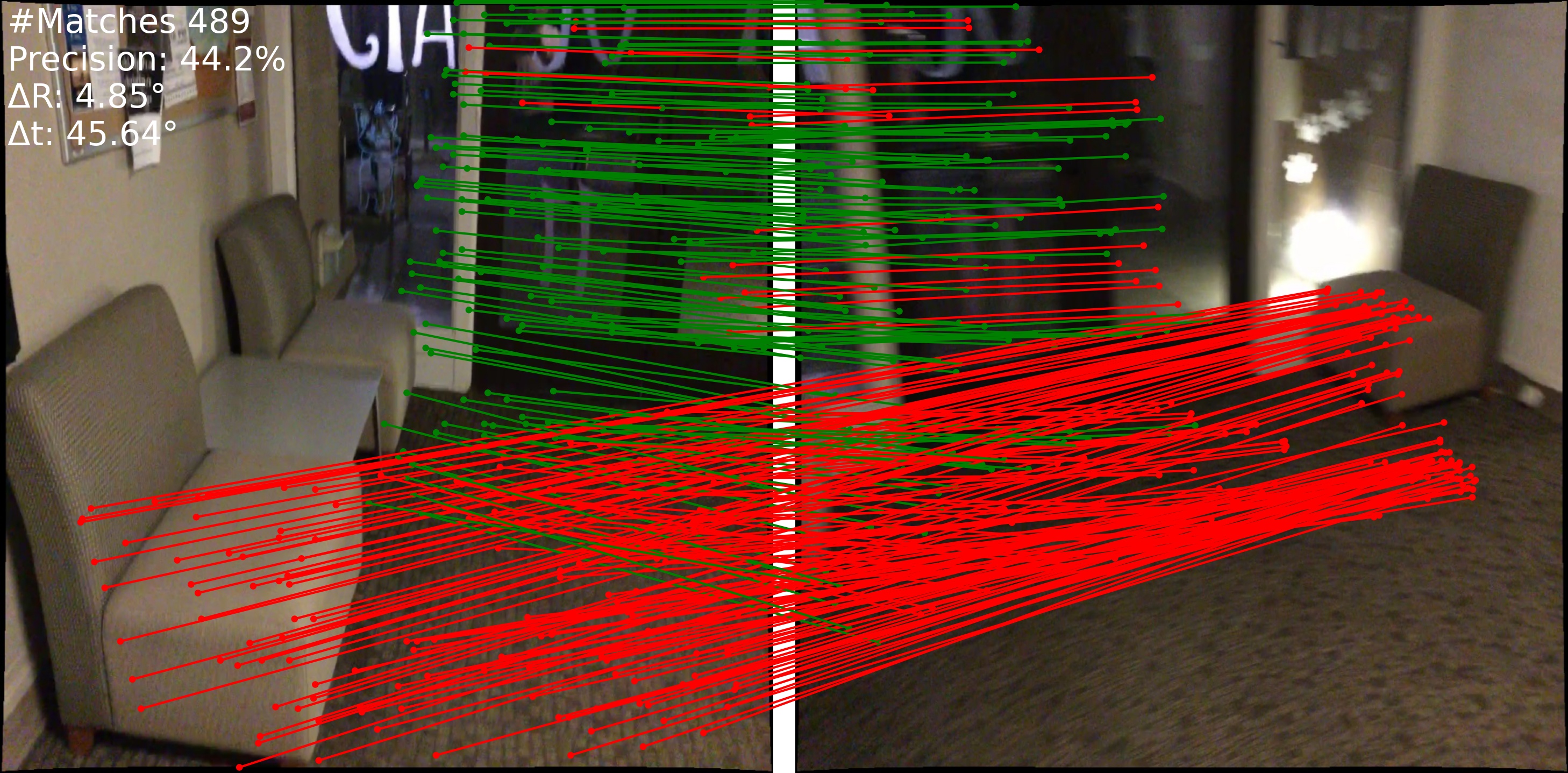}}
		\\
				\raisebox{-0.5\height}{\rotatebox{90}{Outdoor}} &
\raisebox{-0.5\height}{\includegraphics[width=0.3\linewidth]{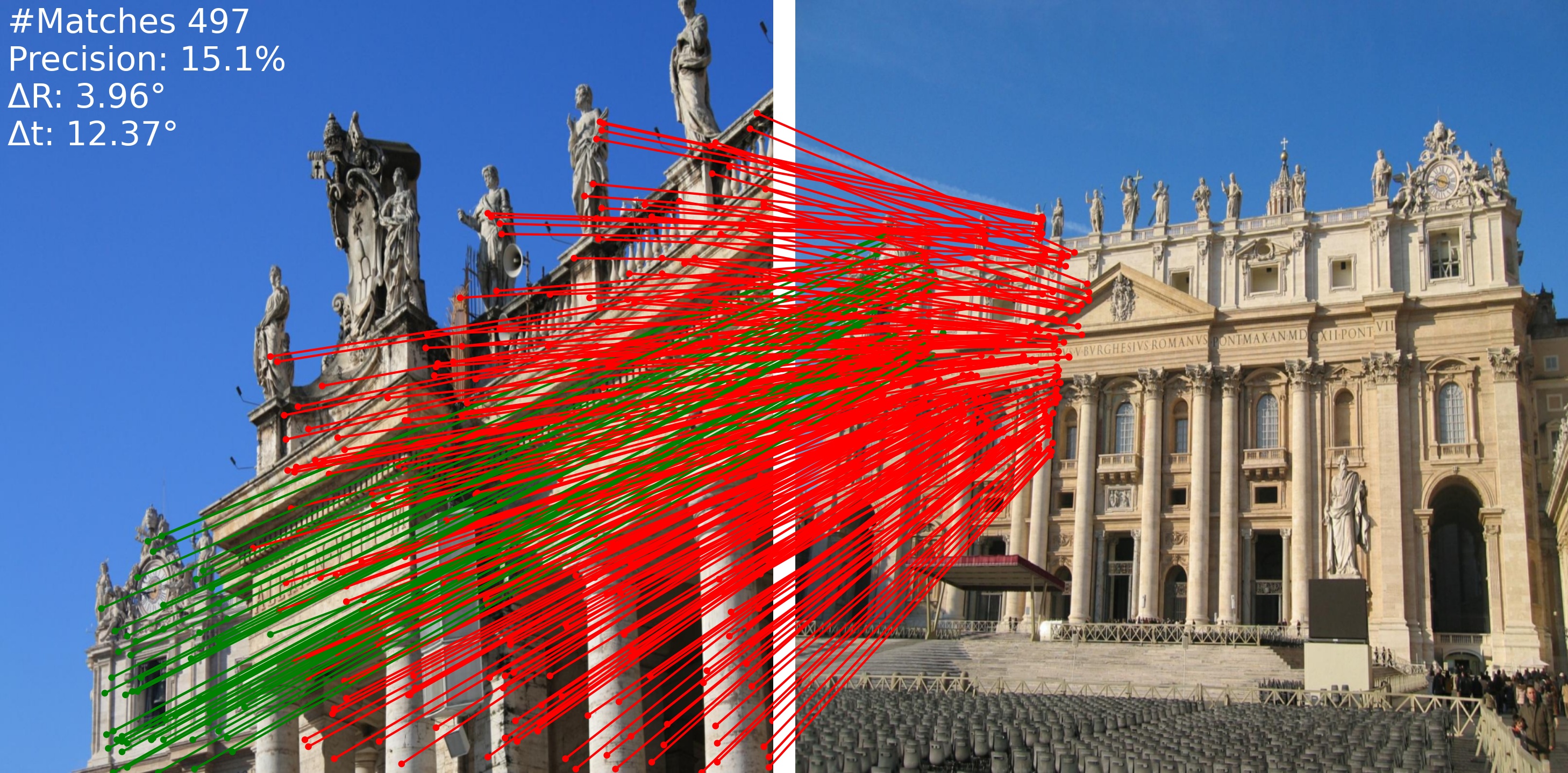}}
\raisebox{-0.5\height}{\includegraphics[width=0.3\linewidth]{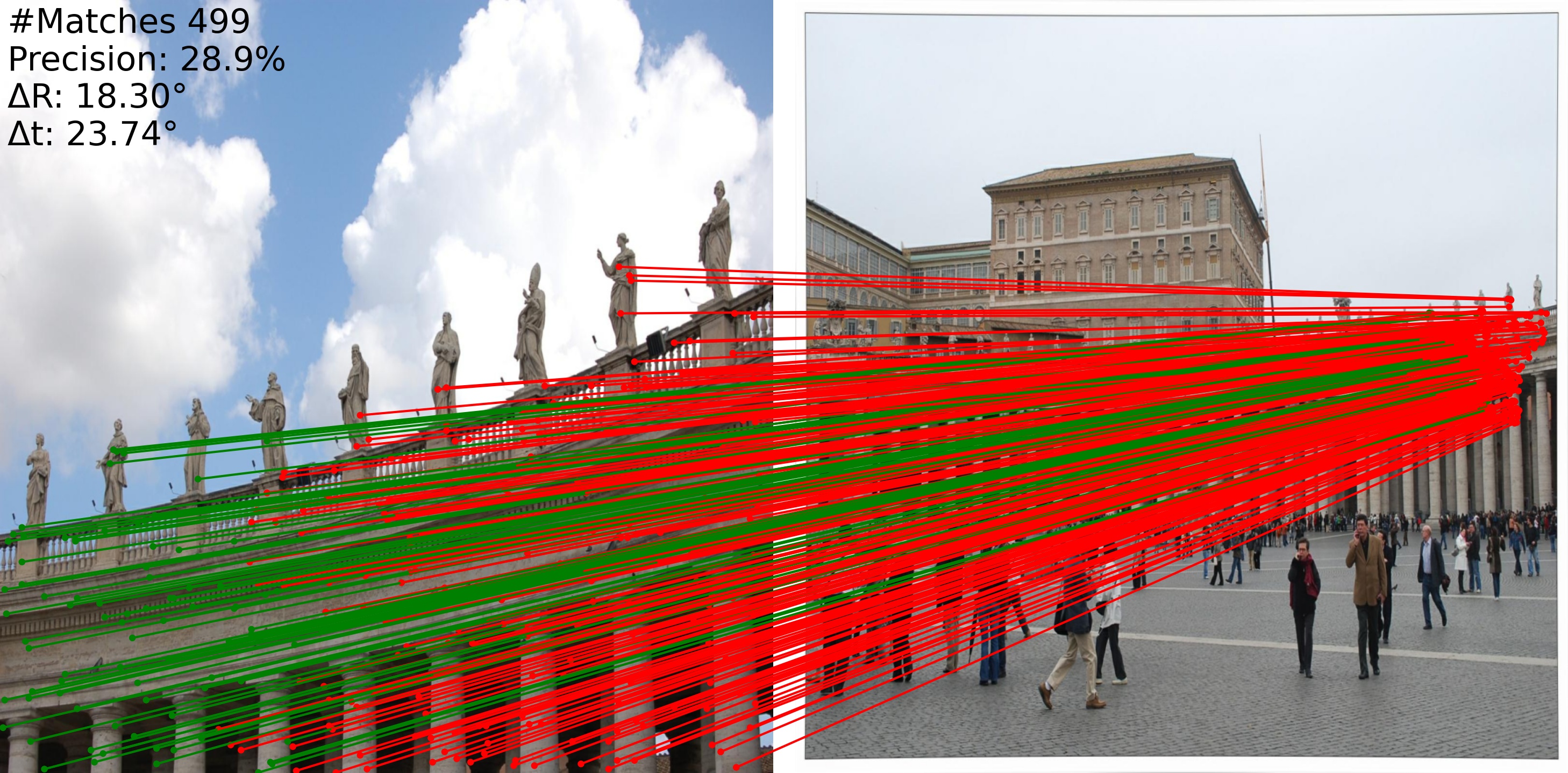}}
\raisebox{-0.5\height}{\includegraphics[width=0.3\linewidth]{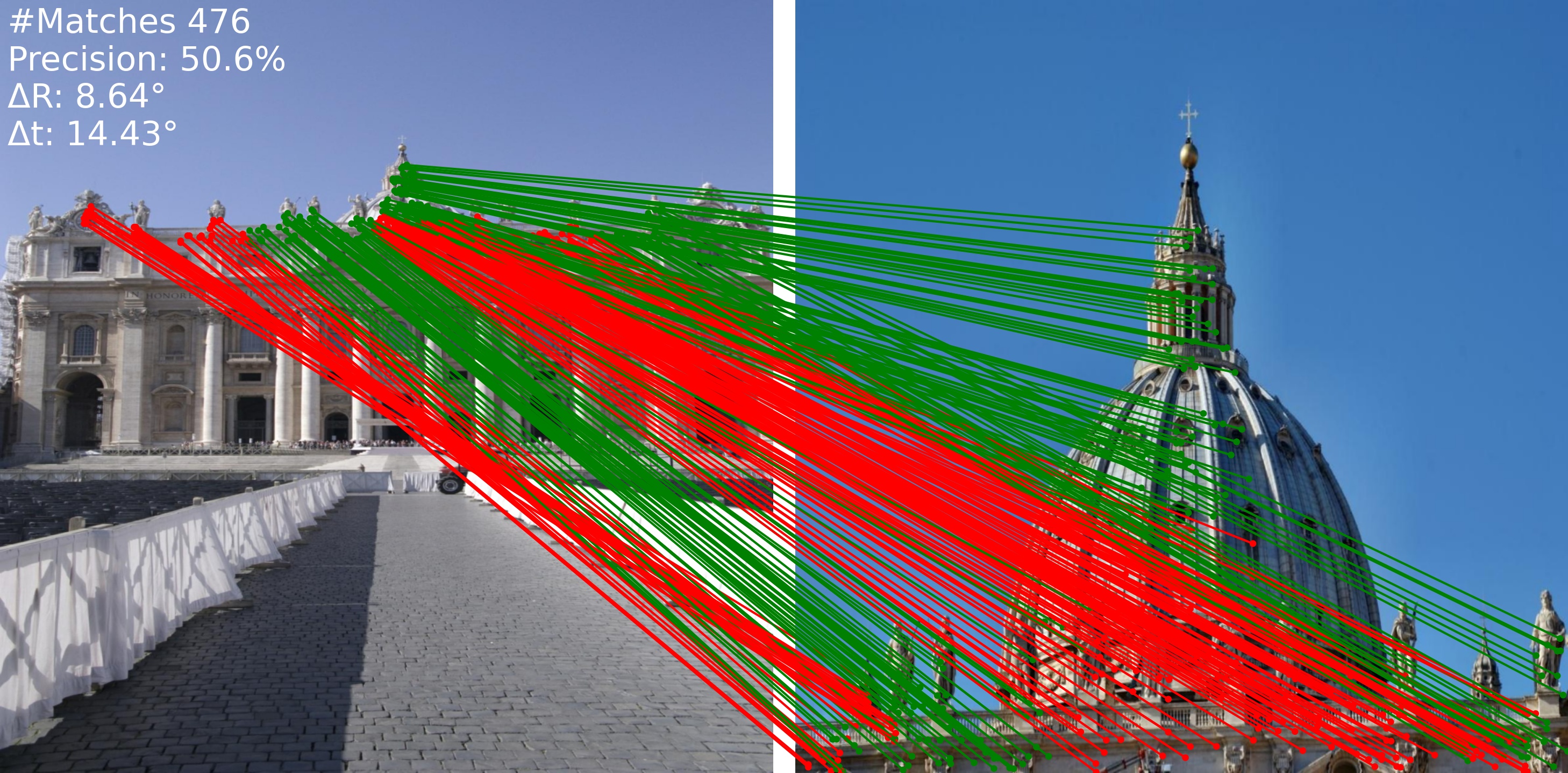}}
	\end{tabular}

	\end{center}
\caption{\textbf{Failure cases on indoor and outdoor data.} Red denotes correspondences with an epipolar error greater than $5 \times 10^{-4}$ in normalized image coordinates.}
\label{fig_failure}

\end{figure}

\textbf{Failure Cases.}
As shown in Fig.~\ref{fig_failure}, CoReMatch's failure cases typically occur under large viewpoint changes combined with textureless regions or repetitive and visually similar structures, where reliable correspondence becomes inherently ambiguous.

\section{Related Work}

\subsection{Learning-Based Feature Matching}

Classical feature matching pipelines typically consist of keypoint detection, descriptor extraction, and descriptor matching using handcrafted representations such as SIFT~\cite{lowe2004distinctive} and ORB~\cite{rublee2011orb}. While computationally efficient, these methods often struggle under large viewpoint changes, weak texture, and significant illumination variations. Learning-based approaches have substantially improved robustness by learning discriminative visual representations and leveraging global contextual reasoning. Early deep matching methods primarily focused on learning local descriptors or improving keypoint repeatability. SuperPoint~\cite{detone2018superpoint} introduced self-supervised keypoint detection and description, while SuperGlue~\cite{sarlin2020superglue} incorporated graph neural networks and attention-based context aggregation for robust feature matching. LightGlue~\cite{lindenberger2023lightglue} further improved efficiency through adaptive Transformer inference.

More recently, semi-dense feature matching methods have gained popularity by eliminating explicit keypoint detection and directly establishing correspondences from dense feature maps. LoFTR~\cite{sun2021loftr} introduced Transformer~\cite{vaswani2017attention}-based coarse matching followed by local fine refinement, achieving strong performance in low-texture regions. Subsequent works have improved efficiency and robustness through adaptive sparse attention and hierarchical refinement. QuadTree~\cite{tang2022quadtree} hierarchically restricts attention to spatially relevant regions, while TopicFM~\cite{giang2023topicfm} limits attention by grouping semantically similar features. ASpanFormer~\cite{chen2022aspanformer} introduced adaptive span attention for efficient long-range reasoning, while Efficient LoFTR~\cite{wang2024eloftr} introduced Aggregated Attention Modules (AAM) to reduce Transformer complexity on high-resolution inputs. EDM~\cite{Li_2025_ICCV} further improves efficiency using a lightweight CNN with correlation injection for multi-scale feature fusion.

\subsection{Dense Correspondence Estimation}

Dense correspondence estimation has been extensively studied in optical flow and dense image matching~\cite{edstedt2023dkm, edstedt2024roma,wang2026waft,teed2020raft,wang2026swift,ranftl2021vision}. RAFT~\cite{teed2020raft} introduced recurrent all-pairs field transforms, iteratively refining dense flow fields through correlation volumes and recurrent updates. Subsequent methods have extended this paradigm toward higher-resolution and more memory-efficient correspondence estimation. More recently, RoMa~\cite{edstedt2024roma} combined global Transformer-based matching with local correlation refinement and multi-scale warping, achieving strong dense matching performance.
Several recent methods~\cite{wang2026waft,wang2026swift} have sought to reduce reliance on explicit correlation volumes. In particular, WAFT~\cite{wang2026waft} introduced Warping-Alone Field Transforms, demonstrating that iterative optical flow refinement can achieve competitive performance using warping-based updates alone. SWIFT~\cite{wang2026swift} further improves efficiency through a scale-specialized warping-based update mechanism, reducing memory and computational requirements compared to WAFT.

\end{document}